\documentclass[journal,article,submit,pdftex,moreauthors]{Definitions/mdpi} 
\firstpage{1} 
\pubvolume{1}
\issuenum{1}
\articlenumber{0}
\pubyear{2025}
\copyrightyear{2025}

\usepackage{longtable}
\usepackage{array}
\usepackage{booktabs}
\usepackage{adjustbox}
\usepackage{multirow}
\usepackage{geometry}
\usepackage{placeins}
\usepackage{tabularx}
\usepackage{cuted}      
\usepackage{stfloats}   
\usepackage{changepage} 
\usepackage{subcaption}
\usepackage[skip=0pt,font=small]{caption}
\usepackage{tabularray}
\newcolumntype{M}[1]{>{\centering\arraybackslash}m{#1}} 

\datereceived{19 June 2025} 
\dateaccepted{25 June 2025} 
\datepublished{30 June 2025} 
\hreflink{https://doi.org/10.3390/xxxxx} 

\providecommand{\linenumbers}{}

\Title{Causality and ''In-the-Wild'' Video-Based Person Re-ID: A Survey}

\TitleCitation{Title}

\Author{
Md Rashidunnabi $^{1,3,\dagger}$\orcidA{}, 
Kailash Hambarde $^{2}$\orcidB{}, 
Hugo Proença $^{1,2}$*\orcidC{}
}

\AuthorNames{Md Rashidunnabi, Kailash Hambarde, and Hugo Proença}

\isAPAStyle{%
    \AuthorCitation{Rashidunnabi, M., Hambarde, K., \& Proença, H.}
}{%
    \isChicagoStyle{%
        \AuthorCitation{Rashidunnabi, Md, Kailash Hambarde, and Hugo Proença.}
    }{
        \AuthorCitation{Rashidunnabi, M.; Hambarde, K.; Proença, H.}
    }
}

\address{$^{1}$ \quad  Intelligent System Labratory, Department of Computer Science, University of Beira Interior, Covilhã, Portugal\\
$^{2}$ \quad  Instituto de Telecomunicações, University of Beira Interior, Covilhã, Portugal\\
$^{3}$ \quad DeepNeuronic, Covilhã, Portugal\\
}

\abstract{
Video-based person re-identification (Re-ID) remains brittle in real-world deployments, despite impressive benchmark performance. Most existing models rely on superficial correlations—such as clothing, background, or lighting—that fail to generalize across domains, viewpoints, and temporal variations. This survey examines the emerging role of causal reasoning as a principled alternative to traditional correlation-based approaches in video-based Re-ID. We provide a structured and critical analysis of methods that leverage Structural Causal Models (SCMs), interventions, and counterfactual reasoning to isolate identity-specific features from confounding factors. The survey is organized around a novel taxonomy of causal Re-ID methods, spanning generative disentanglement, domain-invariant modeling, and causal transformers. We review current evaluation metrics and introduce causal-specific robustness measures. In addition, we assess the practical challenges—scalability, fairness, interpretability, and privacy—that must be addressed for real-world adoption. Finally, we identify open problems and outline future research directions that integrate causal modeling with efficient architectures and self-supervised learning. This survey aims to establish a coherent foundation for causal video-based person Re-ID and to catalyze the next phase of research in this rapidly evolving domain.}

\begin{document}

\section{\textbf{Introduction}}\label{sec:introduction}

Video-based person re-identification (Re-ID) is a critical task in computer vision, with applications in surveillance, smart cities, and forensics~\cite{Wu2018}. Unlike image-based Re-ID, which relies on static appearance cues, video-based methods leverage temporal sequences—capturing motion, gait, and behavioral dynamics—to match individuals across non-overlapping camera views~\cite{Geng2024, Xu2017}. This added temporal dimension provides richer identity signals, particularly in unconstrained environments where single-frame models often fail.

Despite substantial progress, most video-based Re-ID systems remain brittle under real-world conditions. Benchmark-leading models degrade sharply when exposed to domain shifts, occlusions, lighting changes, or clothing variations~\cite{Geirhos2020, Zhang2021, Liu2021}. The root cause is methodological: these models are correlation-driven, trained to optimize performance on tightly curated datasets by exploiting superficial cues—such as clothing color or background texture—that do not generalize to real deployments~\cite{Yang2024, Gu2020, Liao2018}. This leads to fragile identity representations that collapse under distribution shift~\cite{Jin2020}.

To overcome these limitations, causal inference offers a fundamentally different paradigm. Rather than modeling statistical associations between visual input and identity, causal methods aim to isolate the \textbf{true generative factors of identity}—such as body shape, gait, and motion patterns—while explicitly controlling for confounding variables like clothing, background, and viewpoint~\cite{Pearl2009, Peters2017, Schlkopf2021}. Structural Causal Models (SCMs), counterfactual reasoning, and interventional training frameworks provide the tools to enforce this separation, enabling models that are more robust, generalizable, and interpretable~\cite{Zhang2021, Ilse2021, Yuan2024, Bareinboim2022}.

\begin{figure}[H]
    \vspace{2pt}  
    \centering
    \includegraphics[width=\linewidth, keepaspectratio]{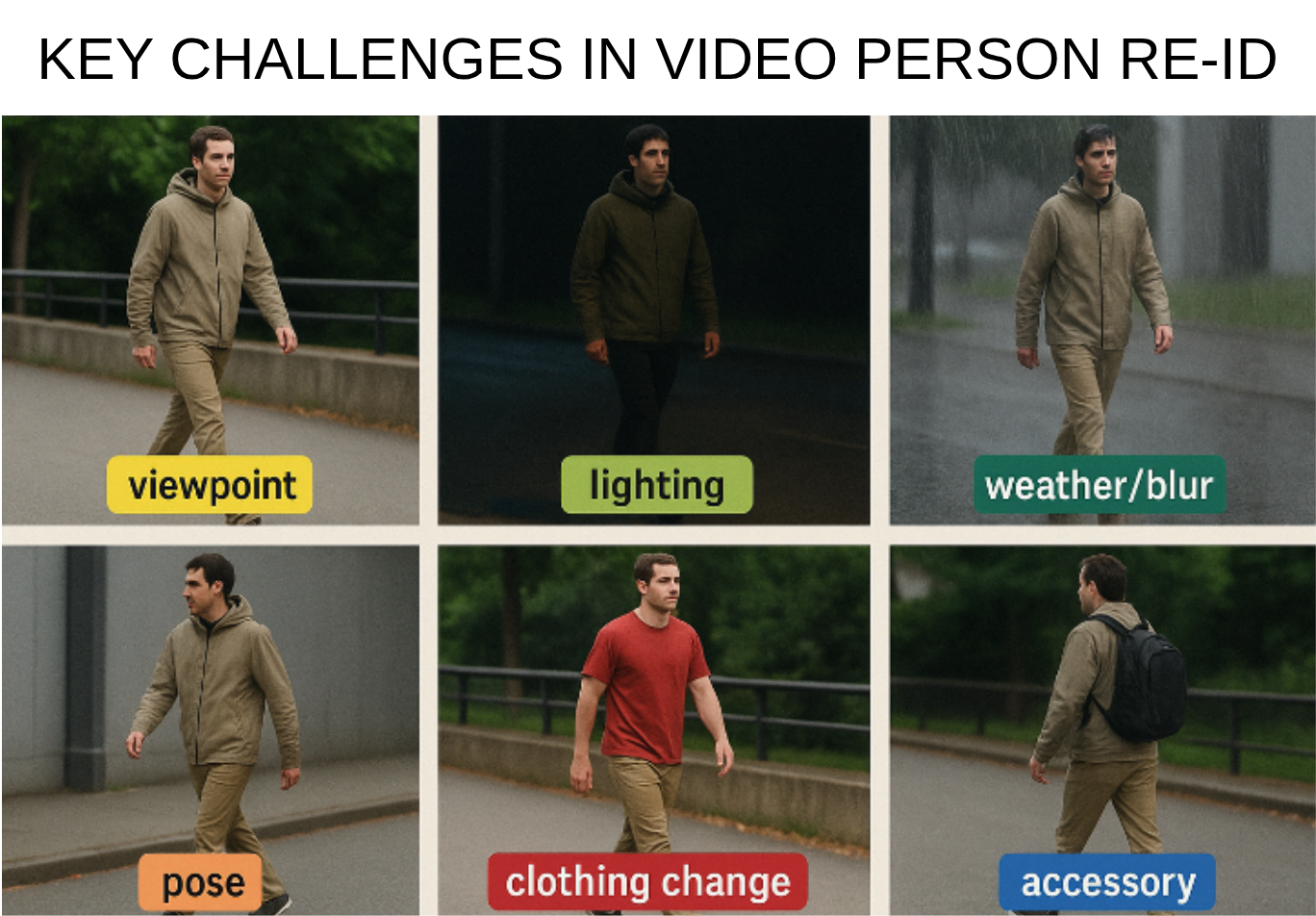}
    \caption{\textbf{Why video-based person Re-ID is hard.} The same individual appears under six nuisance factors—viewpoint, lighting, rain blur, pose, clothing change, and accessory occlusion—illustrating the need for causal disentanglement rather than correlation-driven learning.}
    \label{fig:intro_nuisance}
\end{figure}
\vspace{6pt}  

As illustrated in Figure~\ref{fig:intro_nuisance}, a robust Re-ID system must ignore nuisance variation and preserve consistent identity representations across dramatic appearance shifts. Causal methods explicitly model this requirement by intervening on non-identity attributes and learning representations invariant to them. This shift enables models to resist shortcut learning and to focus on stable identity features that remain consistent across environments.

This survey provides a structured and critical overview of causal approaches in video-based person Re-ID. Our contributions are:

\begin{itemize}
    \item We provide a comprehensive taxonomy of causal methods in Re-ID, covering structural modeling, interventional training, adversarial disentanglement, and counterfactual evaluation;
    \item We review state-of-the-art causal Re-ID models (e.g., DIR-ReID, IS-GAN, UCT) and analyze their performance across real-world challenges such as clothing change, domain shift, and multi-modality;
    \item We propose a unified causal framework for reasoning about identity, confounders, and interventions in Re-ID pipelines;
    \item We discuss emerging causal evaluation metrics, interpretability tools, and benchmark gaps that must be addressed for widespread adoption;
    \item We identify open problems and outline future research directions at the intersection of causality, efficiency, privacy, and fairness in real-world Re-ID systems.
\end{itemize}
\begin{figure}[H]
    \vspace{2pt}
    \centering
    \includegraphics[width=0.9\linewidth]{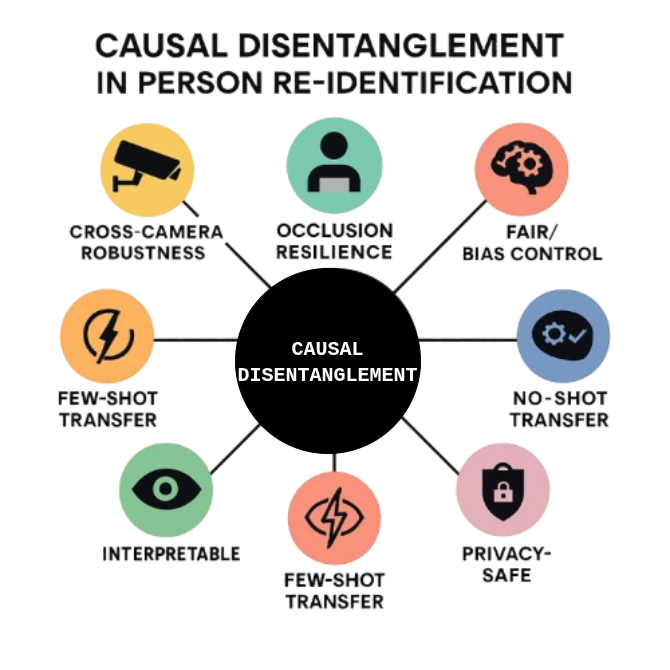}
    \caption{\textbf{Benefits of Causal Disentanglement in Video-Based Person Re-ID.} Causal reasoning improves cross-domain robustness, occlusion resilience, fairness, privacy, and interpretability—key for real-world Re-ID systems.}
    \label{fig:causal_benefits}
\end{figure}
\vspace{6pt}

The remainder of the paper is structured as follows: Section~\ref{sec:fundamentals} reviews the foundations and limitations of conventional Re-ID approaches. Section~\ref{sec:causal} introduces causal inference and formalizes its role in disentangling identity from confounders. Section~\ref{sec:sota} surveys state-of-the-art causal Re-ID models. Section~\ref{sec:disentanglement} details causal disentanglement strategies in practice. Section~\ref{sec:challenges} outlines key open challenges. Section~\ref{sec:future} presents future directions. Finally, Section~\ref{sec:conclusion} summarizes key insights and calls for a paradigm shift from correlation to causation in video-based person Re-ID.

\section{\textbf{Fundamentals of Person Re-Identification}}\label{sec:fundamentals}

\subsection{\textbf{Overview of Video-Based Person Re-ID}}\label{sec:overview-reid}

Video-based person re-identification (Re-ID) focuses on matching individuals across different camera views using sequences of video frames, called tracklets. Unlike single-image Re-ID, which relies on appearance cues, video-based methods leverage both spatial (appearance) and temporal (motion) information. This combination is crucial for distinguishing individuals, especially when appearance alone is unreliable due to variations in viewpoint, illumination, or clothing \cite{Geirhos2020, Geng2024}. Motion information, such as gait and temporal dynamics, plays a significant role in video-based Re-ID. While appearance-based features like clothing color or body shape are useful, they can change due to factors like lighting, posture, or occlusion. In contrast, motion patterns remain relatively stable and can help maintain identity consistency across camera views.

\begin{figure}[htbp]                
    \centering
    \includegraphics[width=.95\linewidth]{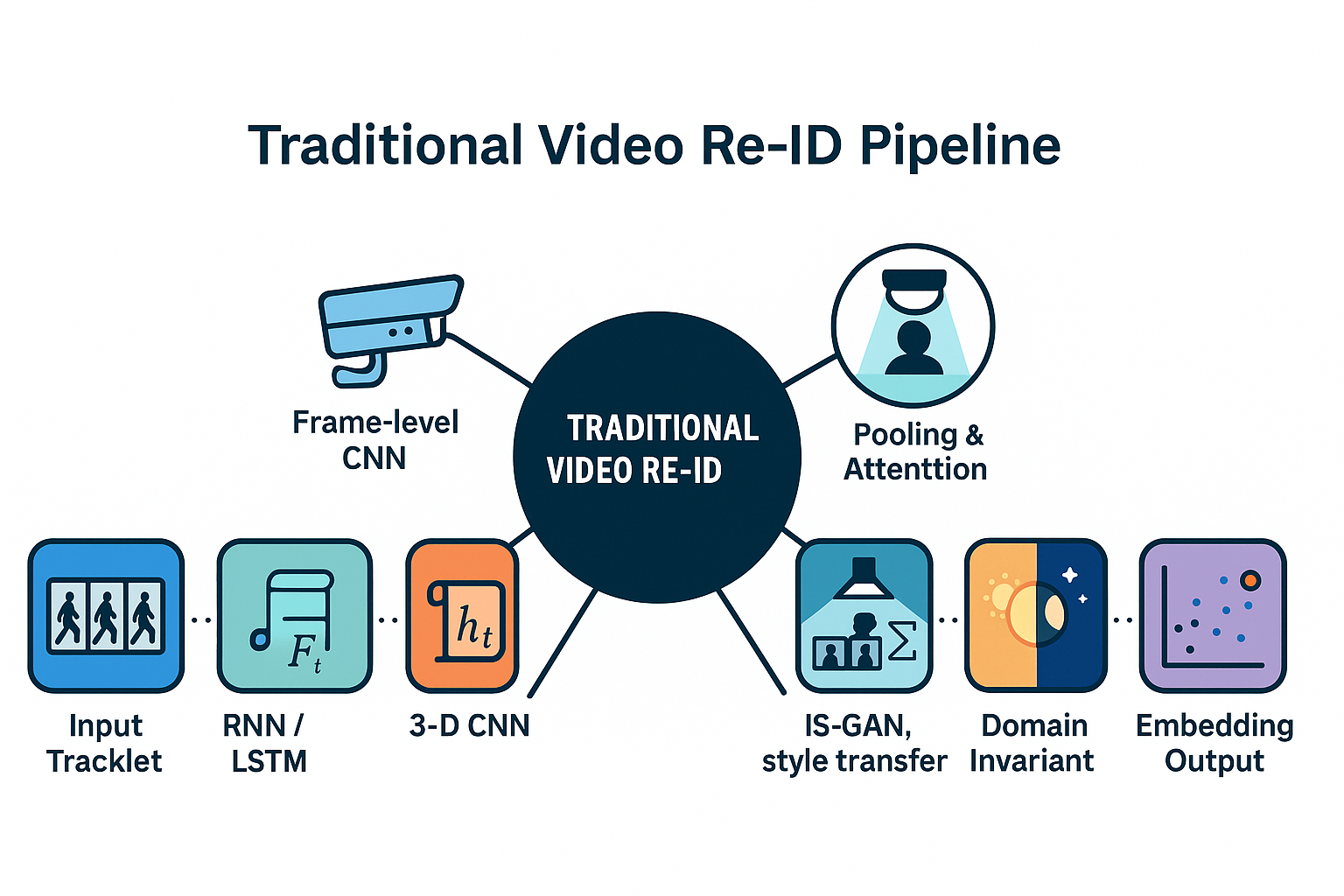}
    \caption{\textbf{Traditional video-based person Re-ID pipeline.}
             The diagram summarises classical modules—frame-level CNN,
             temporal modelling (RNN / 3-D CNN), pooling–attention,
             generative augmentation, and domain-invariant learning—
             that transform a tracklet into a fixed-length identity
             embedding.}
    \label{fig:traditional_pipeline}
  \end{figure}

The typical video-based person Re-ID pipeline, as shown in Figure~\ref{fig:traditional_pipeline}, involves frame-level feature extraction, temporal modeling using RNNs or 3D CNNs, and sequence aggregation to generate a fixed-length identity representation. These methods allow for the capture of both appearance and motion features, which is essential for matching tracklets across non-overlapping camera views \cite{Xu2017, Li2018, Eom2021}.

\subsection{\textbf{Challenges in Video-Based Re-ID}}\label{sec:challenges-reid}

Video-based person re-identification (Re-ID) introduces several complexities compared to single-image person Re-ID due to the dynamic nature of video data. Key challenges include:

\textbf{Occlusions.} In video sequences, individuals are often partially obscured by other objects or people, causing missing identity features. These occlusions can significantly hinder the model's ability to match tracklets across non-overlapping camera views, leading to errors in identity classification.

\textbf{Viewpoint Variations.} Viewpoint changes occur when individuals are captured by cameras positioned at different angles. This results in variations in appearance, as features like body shape and face may look different from different viewpoints. Video-based methods need to account for these changes, typically by utilizing temporal information such as gait and motion patterns, which remain stable across camera views \cite{Geirhos2020, Zhang2021}.

\textbf{Lighting Variations.} Lighting shifts, such as day-to-night or artificial lighting changes, can cause significant color and texture changes in appearance. This can distort visual features like clothing or skin tone, leading to performance degradation in traditional appearance-based methods. Temporal modeling and domain-invariant learning techniques help mitigate these lighting-induced discrepancies \cite{Yang2024, Zhang2021}.

\textbf{Environmental Factors.} Additional environmental factors, such as weather conditions (e.g., rain or fog), background clutter, and scene distractions, can introduce noise into the feature extraction process, further complicating identity matching. video-based person Re-ID systems must be robust to these variations, isolating true identity features from contextual distractions \cite{Geirhos2020, Zhang2021}.

These challenges—occlusions, viewpoint variations, lighting changes, and environmental factors—necessitate video-based person Re-ID systems that can robustly handle dynamic conditions. Models must be designed to focus on identity-specific features and incorporate temporal information to account for these complexities.

\subsection{\textbf{Traditional Approaches and Their Limitations}}\label{sec:traditional-approaches}

Traditional video-based person re-identification (Re-ID) methods rely on several techniques, starting with \textbf{frame-level feature extraction}, where Convolutional Neural Networks (CNNs) such as ResNet-50 are used to extract appearance features from individual video frames \cite{Geng2024}. The frame-level features \( F_t \) for the \( t \)-th frame are computed as \( F_t = \text{CNN}(x_t; \theta_{\text{cnn}}) \), where \( x_t \) is the input frame and \( \theta_{\text{cnn}} \) are the learned parameters of the CNN. These features are then processed through \textbf{temporal modeling} using \textbf{Recurrent Neural Networks (RNNs)} or \textbf{3D Convolutional Neural Networks (CNNs)} to capture temporal dependencies across the frames \cite{Gao2018, Xu2017}. For example, in RNNs, the hidden state \( h_t \) is updated as \( h_t = g(W_h h_{t-1} + W_x F_t + b_h) \), where \( g(\cdot) \) is the activation function, and \( W_h, W_x \) are the weight matrices. After temporal modeling, \textbf{sequence aggregation} methods like \textbf{pooling} or \textbf{attention-based aggregation} are used to combine the frame-level features into a fixed-length identity embedding \( F_{\text{tracklet}} \) for matching across different camera views \cite{Xu2017, Li2018}. The attention mechanism assigns an importance weight \( \alpha_t = \frac{\exp(W_a h_t)}{\sum_{t=1}^{T} \exp(W_a h_t)} \) to each frame based on its relevance. Additionally, generative models such as \textbf{Identity Shuffle GAN (IS-GAN)} and \textbf{Domain-Invariant Representation Learning (DIR-ReID)} help improve generalization across domain shifts by disentangling identity features from background noise \cite{Eom2021, Zhang2021}, making these traditional approaches more robust in real-world scenarios where variations in environment, pose, and lighting can disrupt performance.

However, these traditional approaches exhibit several critical limitations. Despite their ability to leverage both appearance and motion information, they still struggle to generalize under real-world conditions. One primary issue is their reliance on correlations between identity features and non-identity cues, such as background, clothing, and camera angle. This causes the models to overfit to contextual information rather than focusing on identity-specific features, leading to poor performance when domain shifts occur, such as clothing changes, different lighting conditions, or varying camera angles \cite{Yang2024, Geirhos2020}. For example, appearance-based features, such as clothing color, can be misleading due to clothing variations or lighting changes \cite{Tian2018, Zhang2021}. Temporal models like RNNs, although capable of capturing motion, are often ineffective at handling occlusions or partial visibility, which is common in real-world scenarios, as demonstrated by the lack of robustness in tracking over long sequences or through occluded frames \cite{Gao2018, Jia2021}. Additionally, these methods often fail to distinguish between identity cues and environmental factors, making them sensitive to irrelevant noise in the data.

Moreover, while domain-invariant methods like DIR-ReID attempt to address some of these limitations, they still rely heavily on statistical correlations that can easily be confounded by spurious factors, rather than truly isolating identity-specific information from non-identity factors \cite{Zhang2021, Eom2021}. For instance, these models may still rely on camera-specific biases or other nuisance factors that do not correspond to intrinsic identity features, reducing their robustness across different environments or settings. As a result, traditional approaches often underperform when deployed in real-world surveillance applications where conditions are highly variable.

The fundamental limitation of traditional video-based person Re-ID approaches can be summarized as a conceptual reliance on correlation rather than causation~\cite{Zhang2021, Pearl2009}. While correlation-based methods can identify statistical patterns between inputs and identity labels, they cannot distinguish whether these patterns represent genuine identity characteristics or merely coincidental associations~\cite{Geirhos2020, Locatello2019}. This represents a critical gap that necessitates a paradigm shift in how video-based person Re-ID systems approach the identity matching problem~\cite{Yang2024, Subramaniam2019}. Causal modeling provides this shift by formalizing the distinction between identity-relevant factors (like gait patterns and body structure) and identity-irrelevant factors (like background scenes or lighting conditions)~\cite{Schlkopf2021, Peters2017}. By explicitly modeling these causal relationships rather than statistical correlations, causal methods can achieve what traditional approaches fundamentally cannot: reliable identity matching across dramatic environmental variations~\cite{Wang2022, Wu2018}.

Causal inference methods offer a promising solution to these issues by focusing on isolating true identity signals from spurious context. Structural Causal Models (SCMs) and counterfactual reasoning allow for explicit modeling of the relationships between identity-specific factors and non-identity factors like clothing, background, and lighting \cite{Pearl2009, Zhang2021}. These causal methods aim to identify the true causal influence of identity on visual features and remove the influence of confounding variables, such as background noise or clothing changes, through intervention and backdoor adjustment \cite{Zhang2021, Yuan2024}. By explicitly separating identity-related factors from environmental influences, causal models can maintain robust identity representations even under conditions of occlusion, lighting variations, or domain shifts. This shift from purely correlation-based learning to causal inference-based methods promises to enhance generalization, improve robustness, and provide more reliable performance in dynamic real-world environments \cite{Zhang2021, Eom2021}.

\subsection{\textbf{The Role of Visual Attributes in Video-based Person Re-Identification}}

Visual attributes such as clothing color, body shape, gait, and texture are essential in video-based person re-identification (Re-ID) as they bridge low-level pixel data and high-level identity features. These attributes provide human-interpretable cues, improving video-based person Re-ID robustness in challenging scenarios like occlusions, pose variations, and domain shifts~\cite{Tian2018}.

\textbf{Attribute-Based Disentanglement} is key to isolating identity-specific features from non-identity variations like background clutter and clothing changes. Techniques like the Identity Shuffle GAN (IS-GAN)~\cite{Eom2021} factorize images into identity-related and non-identity features, enhancing model generalization. \textbf{Frequency-based Extraction} using 3D Discrete Cosine Transform (3D DCT)~\cite{Liu2023} isolates discriminative patterns, while \textbf{Causal-Based Disentanglement} with Structural Causal Models (SCMs)~\cite{Zhang2021} removes domain-specific biases, improving cross-domain generalization.

\textbf{Occlusion-Resilient Learning}, like that in DRL-Net~\cite{Jia2021}, uses transformer-based models to disentangle visible attributes from occlusions, ensuring accurate identity matching despite partial visibility.

\textbf{Matching and Filtering} based on attribute similarity helps refine identity matches, while \textbf{Interpretability} benefits from attribute-based models like ASA-Net, which clarifies decision-making~\cite{Tian2018}. However, \textbf{Bias and Fairness} concerns arise as attributes like gender and age may introduce discrimination if not handled carefully.

\begin{table}[H]
\caption{Common semantic attributes in video-based person Re-ID and representative extraction pipelines.}
\label{tab:sem_attr_reid}
\centering
\begin{tabularx}{\textwidth}{>{\raggedright\arraybackslash}p{3.2cm} >{\centering\arraybackslash}p{2.4cm} X}
\toprule
\textbf{Attribute Type} & \textbf{Static / Dynamic} & \textbf{Typical Extraction Method (key reference)} \\
\midrule
Clothing Colour               & Static   & Colour histograms, Retinex–LOMO descriptor\cite{liao2015lomo}                                           \\
Clothing Category (shirt / pants) & Static   & Part-based CNN multi-task attribute head (APR-Net)\cite{lin2017apr}                                   \\
Accessories (bags, hats, and other accessories)
  & Static   & Weakly-supervised multi-scale attribute localisation\cite{tang2019msasl}; mid-level attribute CNN\cite{zhang2018mlap} \\
Gait / Silhouette             & Dynamic & Set-level silhouette sequence model (GaitSet)\cite{chao2019gaitset}                                    \\
Body Shape / Height           & Static & 3-D skeleton key-point statistics\cite{munaro2014skeleton}                                             \\
Texture / Pattern             & Static   & Local Gaussian / SILTP texture blocks (HGD + LOMO)\cite{matsukawa2016hgd,liao2015lomo}                \\
Gender / Age / Hair           & Static   & Multi-task mid-level attribute + identity CNN\cite{zhang2018mlap}                                     \\
Pose / Motion State           & Dynamic & Pose-driven deep convolutional model with RPN attention\cite{su2017pdc}                                \\
Carried Objects               & Dynamic & Attribute-aware object detectors / semantic parts\cite{zhang2018mlap,tang2019msasl}                   \\
\bottomrule
\end{tabularx}
\end{table}

\subsection{\textbf{Attribute-Specific Evaluation Metrics for Video-Based Person Re-Identification}}

Video-based person Re-ID systems have traditionally used standard metrics like Cumulative Matching Characteristic (CMC) and mean Average Precision (mAP) to evaluate performance~\cite{ZhengICCV2015, Oliveira2021}. However, recent approaches have introduced attribute-specific metrics that capture more nuanced aspects of model behavior, including soft-biometric consistency, occlusion robustness, and causal sensitivity~\cite{KhamisSamehandKuo2015}.

\textbf{Traditional Retrieval Metrics.} CMC measures the probability of finding a correct match within the top-k ranks, defined as \( \text{CMC@}k = \frac{1}{N}\sum_{i=1}^N \mathbf{1}(\text{rank}(i) \le k) \), where \(N\) is the number of queries, and the indicator function \( \mathbf{1}(\cdot) \) returns 1 if the rank is within the top \(k\)~\cite{Hermans2017, Yan2017}. While widely used in benchmarks like MARS and DukeMTMC-VideoReID, CMC is limited by its single-match focus and sensitivity to gallery size. In contrast, mAP captures both precision and recall, defined as \( \text{mAP} = \frac{1}{N} \sum_{q=1}^N \text{AP}(q) \), where Average Precision (AP) represents the area under the precision-recall curve for each query, offering a more comprehensive assessment~\cite{li2019lsvid}.

\textbf{Attribute-Level Metrics.} These metrics evaluate consistency across soft-biometric attributes, including Attribute Consistency, which measures the fraction of matching attributes in retrieved pairs, and Attribute-Aware Accuracy, which conditions retrieval accuracy on attribute agreement~\cite{Chen2020, Chai2021}. Occlusion Robustness assesses accuracy under partial occlusions, while Clothing-Change Robustness evaluates stability across different outfits~\cite{Xu2023, Li2023}. Identity Switch Rate (IDSR) or IDF1, adapted from multi-object tracking, quantifies identity flips across frames, reflecting long-term tracking stability~\cite{Luiten2021}.

\textbf{Causal Robustness Metrics.} Causal-inspired metrics, such as Counterfactual Consistency, test whether identity predictions remain stable under hypothetical attribute changes, probing a model's reliance on true causal signals~\cite{suter2019robustly}. Causal Saliency Ranking ranks features by their causal influence on identity matching, while Intervention-Based Score Shift measures the change in matching scores under controlled attribute interventions, highlighting sensitivity to specific visual cues~\cite{black2021consistent}.

These advanced metrics provide deeper insights into model robustness, interpretability, and generalization, moving beyond simple precision-recall evaluations to capture the complex challenges of real-world Re-ID~\cite{Qian2024, Alkanat2020}.

\unskip
\begin{table}[H] 
\caption{Attribute-Specific Evaluation Metrics for Video-Based Person Re-Identification.}
\label{tab:attribute_metrics}
\begin{adjustwidth}{-\extralength}{0cm}
\begin{tabularx}{\fulllength}{
    >{\raggedright\arraybackslash}m{3.0cm}  
    >{\raggedright\arraybackslash}X  
    >{\raggedright\arraybackslash}X  
    >{\raggedright\arraybackslash}X  
}
\toprule
\textbf{Metric} & \textbf{Measures} & \textbf{Used In / Reports} & \textbf{Advantages / Limitations} \\
\midrule
CMC / Rank-$k$ Accuracy~\cite{Hermans2017} 
& Probability of correct match within rank $k$ (precision at $k$). 
& Almost all Re-ID (image \& video); e.g., MARS~\cite{Zheng2016}, DukeMTMC-VideoReID~\cite{Ristani2016}, SYSU~\cite{Wu2017}. 
& Standard precision metric; lacks recall information. \\
\midrule
Rank-1 Accuracy~\cite{ZhengICCV2015} 
& Top-1 retrieval accuracy ($\text{CMC}@1$). 
& Standard benchmark metric in most Re-ID works~\cite{Wu2023, Li2022}. 
& Single-number summary; no recall information. \\
\midrule
Mean Average Precision (mAP)~\cite{Li2019} 
& Overall retrieval quality (precision and recall averaged). 
& Used in Re-ID benchmarks (Market-1501, MARS~\cite{Zheng2016}, etc.) 
& Comprehensive metric, but sensitive to outliers. \\
\midrule
Attribute Consistency~\cite{Chen2020} 
& Semantic consistency across views. 
& Attribute-based Re-ID works~\cite{Chai2021, KhamisSamehandKuo2015}. 
& Reveals stable cues, but depends on attribute annotation quality. \\
\midrule
Attribute-Aware Accuracy~\cite{Chen2020} 
& Retrieval accuracy with attribute agreement. 
& Joint attribute/ID methods~\cite{Chai2021, Tian2018}. 
& Fine-grained measure, but rarely reported. \\
\midrule
Occlusion Robustness~\cite{Jia2021} 
& Drop in performance under occlusion. 
& Occluded-Duke, Occluded-REID~\cite{Zhao2020}. 
& Useful for real-world scenarios; needs labeled occlusions. \\
\midrule
Clothing-Change Robustness~\cite{Xu2023} 
& Sensitivity to apparel changes. 
& Long-term Re-ID (e.g., DeepChange~\cite{Xu2023}). 
& Reveals clothing cue reliance; needs paired outfits. \\
\midrule
IDSR / IDF1~\cite{Luiten2021} 
& ID switch frequency. 
& Multi-camera tracking~\cite{Luiten2021, Ristani2016}. 
& Consistency metric; requires track-level GT. \\
\midrule
Counterfactual Consistency~\cite{black2021consistent} 
& Invariance to manipulated attributes. 
& Emerging causal Re-ID metrics~\cite{Yang2024, Rao2021}. 
& Tests reliance on stable ID features; challenging to implement. \\
\midrule
Causal Saliency Ranking~\cite{Yuan2024} 
& Importance of features for ID matches. 
& Explainable Re-ID studies~\cite{Li2023, Zhang2021}. 
& Reveals true causal drivers, but lacks numeric comparability. \\
\midrule
Intervention-Based Score Shift~\cite{Schlkopf2021} 
& Effect of controlled attribute interventions. 
& Causal evaluation studies~\cite{Yuan2024, Sun2023}. 
& Quantifies sensitivity; requires well-defined interventions. \\
\bottomrule
\end{tabularx}
\end{adjustwidth}
\end{table}

Table~\ref{tab:attribute_metrics} summarizes a range of evaluation metrics for video-based person Re-ID, spanning traditional measures like CMC, Rank-1, and mAP, as well as more specialized, attribute-specific metrics. While CMC and mAP capture overall retrieval accuracy, attribute-level metrics like Attribute Consistency and Attribute-Aware Accuracy focus on maintaining soft-biometric consistency, reflecting the stability of identity features across views. Metrics like Occlusion Robustness and Clothing-Change Robustness assess model resilience to partial occlusions and outfit variations, respectively. Emerging causal metrics, such as Counterfactual Consistency and Causal Saliency Ranking, aim to evaluate the impact of specific attributes on identity prediction, supporting more interpretable and context-aware video-based person Re-ID systems.

\subsection{\textbf{Common Datasets for Video-Based Person Re-Identification}}\label{subsec:datasets-metrics}

Video-based person Re-ID datasets come in several forms, including visible-spectrum, cross-modality, and synthetic datasets. These datasets vary in scale, diversity, and complexity, offering distinct challenges for model evaluation. Table \ref{tab:dataset_summary} provides a comprehensive summary of these datasets, highlighting key attributes such as the number of identities, sequence counts and camera setups.


\unskip
\unskip
\begin{adjustwidth}{-\extralength}{0cm}
\begin{longtblr}[
    caption = {Comparative Summary of Common Datasets for Video-Based Person Re-Identification.},
    label = {tab:dataset_summary}
]{
    colspec = {X[16,l,m] X[7,c,m] X[12,l,m] X[16,l,m] X[25,l,m] X[14,l,m] X[10,c,m]},
    rowhead = 1,
    row{1} = {font=\bfseries},
    hline{1,Z} = {1pt},
    hline{2-Y} = {0.5pt},
    cell{1}{1-7} = {font=\small\bfseries, m},
    cell{2-Z}{1-7} = {font=\small}
}
\SetCell{c} Dataset & \SetCell{c} Year & \SetCell{c} Modality & \SetCell{c} Identities & \SetCell{c} Sequences / Images & \SetCell{c} Cameras & \SetCell{c} Dataset Link \\
PRID2011~\cite{hirzer11} & 2011 & RGB & 934 total (200 overlap) & 385 (camA) + 749 (camB) & 2 & \href{https://www.tugraz.at/institute/icg/research/team-bischof/learning-recognition-surveillance/downloads/prid11}{Download} \\
iLIDS-VID~\cite{Wang2014} & 2014 & RGB & 300 & 600 (300$\times$2) & 2 & \href{https://xiatian-zhu.github.io/downloads_qmul_iLIDS-VID_ReID_dataset.html}{Download} \\
MARS~\cite{Zheng2016} & 2016 & RGB & 1,261 & $\approx$ 20,000 tracklets (incl.\ 3,248 distractors) & 6 & \href{https://link.springer.com/chapter/10.1007/978-3-319-46466-4_52}{Download} \\
SYSU-MM01~\cite{Wu2017} & 2017 & RGB \& Thermal & 491 & 287,628 RGB + 15,729 IR & 6 (4 RGB, 2 IR) & \href{https://github.com/wuancong/SYSU-MM01}{Download} \\
RegDB~\cite{Ye2021} & 2017 & RGB \& Thermal & 412 & 4,120 (10 vis + 10 IR per ID) & 2 (1 vis, 1 IR) & \href{https://opendatalab.com/OpenDataLab/RegDB}{Download} \\
DukeMTMC-VideoReID~\cite{Ristani2016} & 2018 & RGB & 1,404 (702 train + 702 test) + 408 distractors & 4,832 (2,196 train + 2,636 test) & 8 & \href{https://github.com/Yu-Wu/DukeMTMC-VideoReID}{Download} \\
LS-VID~\cite{li2019lsvid} & 2019 & RGB & 3,772 & 14,943 tracks ($\approx$ 3M frames) & 15 (3 indoor, 12 outdoor) & \href{https://www.pkuvmc.com/dataset.html}{Download} \\
L-CAS RGB-D-T~\cite{Cosar2019} & 2019 & RGB \& Depth \& Thermal & Not Specified & $\approx$ 4,000 (rosbags) & 3 (RGB, Depth, Thermal) & \href{https://lcas.lincoln.ac.uk/wp/research/data-sets-software/l-cas-rgb-d-t-re-identification-dataset/}{Download} \\
P-DESTRE~\cite{Kumar2020} & 2020 & RGB & 1,581 & Over 40,000 frames & UAVs & \href{https://www.di.ubi.pt/~hugomcp/PReID/}{Download} \\ 
FGPR~\cite{Yin2020} & 2020 & RGB & 358 & 716 & 6 (2 per color group) & \href{https://www.isee-ai.cn/~yinjiahang/FGPR.html}{Download} \\
PoseTrackReID~\cite{Siv2020} & 2020 & RGB & $\approx$ 5,350 & $\approx$ 7,725 tracks & Unknown & \href{https://github.com/numediart/PoseTReID_DATASET}{Download} \\
RandPerson~\cite{Wang2020} & 2020 & Synthetic RGB & 8,000 & 1,801,816 images & 19 (virtual cams) & \href{https://github.com/VideoObjectSearch/RandPerson}{Download} \\
DeepChange~\cite{Xu2023} & 2022 & RGB & 1,121 & 178,407 frames & 17 & \href{https://github.com/PengBoXiangShang/deepchange}{Download} \\
LLVIP~\cite{Jia2021_LLVIP} & 2022 & RGB \& Thermal & $\approx$ (15,488 pairs) & 30,976 images & 2 (1 RGB, 1 IR) & \href{https://bupt-ai-cz.github.io/LLVIP/}{Download} \\
ClonedPerson~\cite{Wang2022} & 2022 & Synthetic RGB & 5,621 & 887,766 images & 24 (virtual cams) & \href{https://github.com/Yanan-Wang-cs/ClonedPerson}{Download} \\
BUPTCampus~\cite{Du2024} & 2023 & RGB \& Thermal & 3,080 & (RGB-IR tracklets) & 2 (1 RGB, 1 IR) & \href{https://github.com/dyhBUPT/BUPTCampus}{Download} \\
MSA-BUPT~\cite{Zhao2024} & 2024 & RGB & 684 & 2,665 & 9 (6 indoor, 3 outdoor) & \href{https://mcprl.com/html/dataset/msa.html}{Download} \\
GPR+~\cite{Xiang2020} & 2024 & Synthetic RGB & 808 & 475,104 bounding boxes & Unknown & \href{https://jeremyxsc.github.io/GPR/}{Download} \\
G2A-VReID~\cite{Zhang2024} & 2024 & RGB & 2,788 & 185,907 images & Ground surveillance \& UAVs & \href{https://github.com/fhr-l/g2a-vreid}{Download} \\ 
DetReIDX~\cite{Hambarde2025} & 2025 & RGB & 509 & 13 million+ annotations & 7 university campuses (3 continents) & \href{https://www.it.ubi.pt/DetReIDX/}{Download} \\
AG-VPReID~\cite{Nguyen2025} & 2025 & RGB & 6,632 & 32,321 tracklets & Drones (15-120m altitude), CCTV, Wearable cameras & \href{https://www.kaggle.com/competitions/agvpreid25}{Download} \\ 
\end{longtblr}
\end{adjustwidth}

Table~\ref{tab:dataset_summary} This table presents a comprehensive overview of widely used video-based person re-identification datasets, covering various modalities such as RGB, RGB-Thermal, Depth, and Synthetic RGB. It highlights critical characteristics like the number of identities, sequences, and cameras, reflecting the diversity in data scales and environmental conditions. For instance, \textbf{PRID2011} captures moderate occlusions and viewpoint changes with 934 identities, while large-scale datasets like \textbf{MARS} and \textbf{LS-VID} offer millions of frames for deep learning models. Cross-modality datasets like \textbf{SYSU-MM01} and \textbf{RegDB} introduce challenging RGB-Infrared matching, supporting domain adaptation research. Synthetic datasets like \textbf{RandPerson} and \textbf{ClonedPerson} enable domain generalization with extensive identity counts and realistic appearance variations, making them essential for robust model evaluation.

\section{\textbf{Causal Foundations for Person Re-Identification}}\label{sec:causal}

Before delving into the application of causal reasoning to person re-identification, it is essential to clarify the key terminology that forms the foundation of this approach:

\begin{itemize}
    \item \textbf{Causal Inference}: Unlike statistical correlation which merely identifies patterns of association, causal inference aims to understand the underlying cause-and-effect relationships between variables~\cite{Pearl2009, Schlkopf2021}. In Re-ID, this means distinguishing which visual features truly cause identity recognition (e.g., body structure) versus those that merely correlate with identity in specific contexts (e.g., clothing)~\cite{Zhang2021}.
    
    \item \textbf{Structural Causal Models (SCMs)}: Mathematical frameworks that use directed graphs to explicitly represent causal relationships between variables~\cite{Pearl2009, Peters2017}. In these graphs, nodes represent variables (such as identity, clothing, or background), and directed edges represent the causal influence of one variable on another~\cite{Zhang2021, Bareinboim2022}.
    
    \item \textbf{Confounding Variables}: Factors that influence both the cause and effect, potentially creating spurious correlations~\cite{Pearl2009, Glymour2019}. In Re-ID, environmental factors like lighting or camera viewpoint can confound the relationship between identity and visual appearance~\cite{Yang2024, Geirhos2020}.
    
    \item \textbf{Intervention}: The process of actively modifying a variable in a causal system to observe the effect on other variables~\cite{Pearl2009, Peters2017}. In Re-ID, this might involve artificially changing a person's clothing in images while keeping their identity constant~\cite{Eom2021, Yuan2024}.
    
    \item \textbf{Counterfactual Reasoning}: Evaluating what would have happened under conditions different from what actually occurred~\cite{Pearl2009, Schlkopf2021}. For Re-ID, this involves asking questions like "Would the model still identify this person correctly if they were wearing different clothes?"~\cite{Sun2023}.
    
    \item \textbf{Causal Disentanglement}: The process of separating variables that are causally independent from one another in the underlying data generation process~\cite{Locatello2019, Schlkopf2021}. In Re-ID, this means isolating identity-specific features from non-identity factors like background or lighting~\cite{Zhang2021, Eom2021}.
\end{itemize}

These concepts provide the theoretical framework for addressing the limitations of traditional video-based person Re-ID approaches by focusing on the true causal factors that determine identity, rather than relying on potentially misleading correlations.

\subsection{\textbf{Introduction to Causal Inference}}

\begin{figure}[H]
    \vspace{2pt}
    \centering
    \includegraphics[width=0.95\linewidth, height=6.3cm]{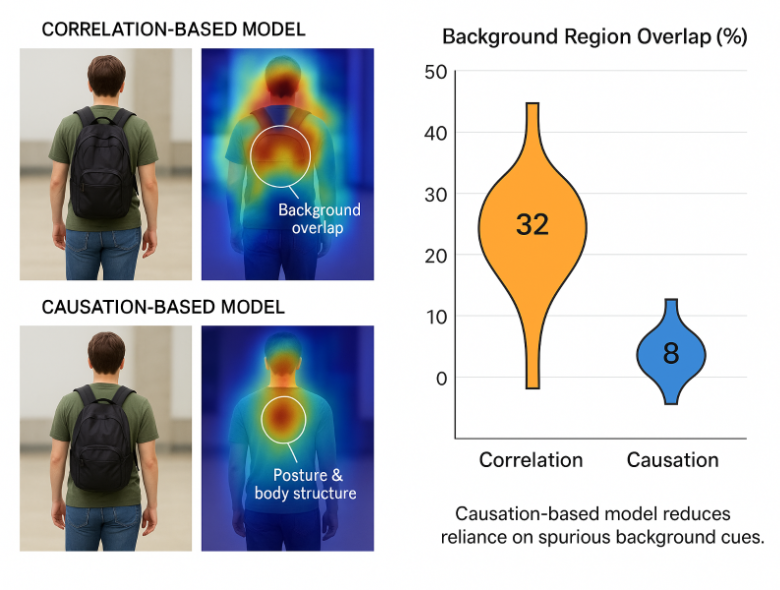}
    \caption{\textbf{Correlation versus causation in Re-ID.} This figure contrasts a correlation based model whose heatmap (top right) overwhelmingly highlights the backpack and surrounding background with a causation based model whose heatmap (bottom right) instead focuses on the upper back neck and head posture as true identity intrinsic features. The illustrative violin plot shows 32\% versus 8\% median background overlap not as experimental values but to emphasize how causal training de-emphasizes spurious context. As discussed in the papers below this shift yields more robust and generalizable person re identification performance than correlation only approaches.}
    \label{fig:correlation_causation}
\end{figure}

Causal inference provides a framework for understanding cause-and-effect relationships in video-based person Re-ID by isolating true identity-preserving features and discarding confounders like background or clothing, which often interfere with traditional models. Unlike correlation-based methods, which rely on spurious associations, causal models focus on identity signals such as body shape, gait, and motion consistency, using causal interventions to remove the influence of confounders like viewpoint or lighting changes. This shift improves video-based person Re-ID performance across domain shifts and environmental variations, as shown in Figure~\ref{fig:correlation_causation}, which contrasts correlation-based and causal models. Structural Causal Models (SCMs) model identity as the cause of observed features, with environmental factors treated as confounders. By applying causal interventions, these models ensure that identity signals remain unaffected by external noise, improving robustness and generalization.

\vspace{6pt}

Causal methods improve video-based person Re-ID accuracy by reducing the influence of confounders that traditional models mistake for identity cues. For instance, while traditional models may incorrectly associate a jacket with identity, causal models maintain accuracy despite changes in appearance due to lighting. DIR-ReID, for example, improves cross-domain Rank-1 accuracy by 11.2\% by removing the causal effect of domain-specific features on appearance~\cite{Zhang2021}. Causal models also excel in handling occlusions by learning the causal relationships between body parts and identity. This allows them to make accurate predictions even when parts of the person are obscured. For example, IS-GAN shows a 15.7\% improvement in Rank-1 accuracy under severe occlusion conditions compared to non-causal models~\cite{Eom2021}. These methods demonstrate how causal inference improves the robustness and reliability of video-based person Re-ID systems in real-world environments.

\subsection{\textbf{Structural Causal Models (SCMs) and Counterfactual Reasoning}}\label{sec:scm_counterfactual}

\begin{figure}[H]
    \vspace{2pt}
    \centering
    \includegraphics[width=0.95\linewidth]{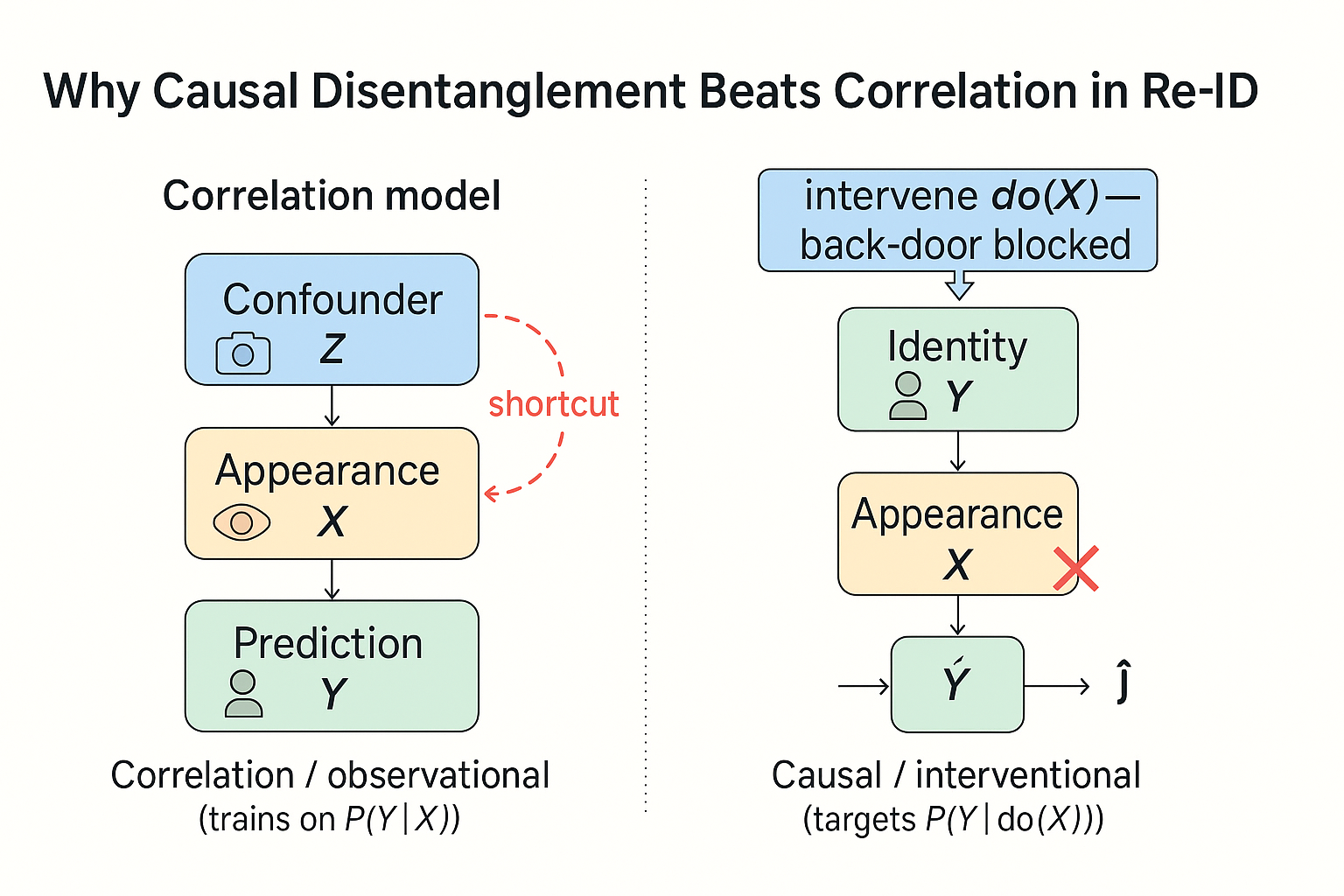}
    \caption{\textbf{Comparing Correlation vs. Causation in Re-ID.} The left side shows the traditional correlation-based approach where confounders ($Z$) can create shortcuts between appearance ($X$) and prediction ($Y$), leading to spurious correlations. The model is trained on observed data $P(Y|X)$, making it vulnerable to changes in the distribution. The right side illustrates the causal/interventional approach that blocks the backdoor path from confounding variables through intervention $do(X)$. By targeting $P(Y|do(X))$, the causal model focuses on the direct effect of identity ($Y$) on appearance ($X$), resulting in more robust predictions under varying conditions.}
    \label{fig:correlation_causation_comparison}
\end{figure}

In video-based person re-identification (Re-ID), \textbf{Structural Causal Models (SCMs)} provide a framework to model the relationships between identity-specific factors and confounders like clothing, background, or camera variations. Unlike traditional models that rely on correlations, SCMs define causal graphs where identity ($I$) influences appearance ($X$), while non-identity factors such as clothing ($C$) and background ($B$) act as confounders~\cite{Pearl2009}. The data generation process can be expressed as $X = f(I, C, B, \text{Camera})$, where the goal is to intervene on non-identity factors and observe how these changes affect identity predictions, thereby isolating the impact of identity itself~\cite{Zhang2021}. Figure~\ref{fig:correlation_causation_comparison} illustrates the fundamental difference between correlation-based and causal approaches to Re-ID, highlighting how causal models block the influence of confounding variables through intervention.

\vspace{6pt}

\begin{figure}[H]
    \vspace{2pt}
    \centering
    \includegraphics[width=\textwidth]{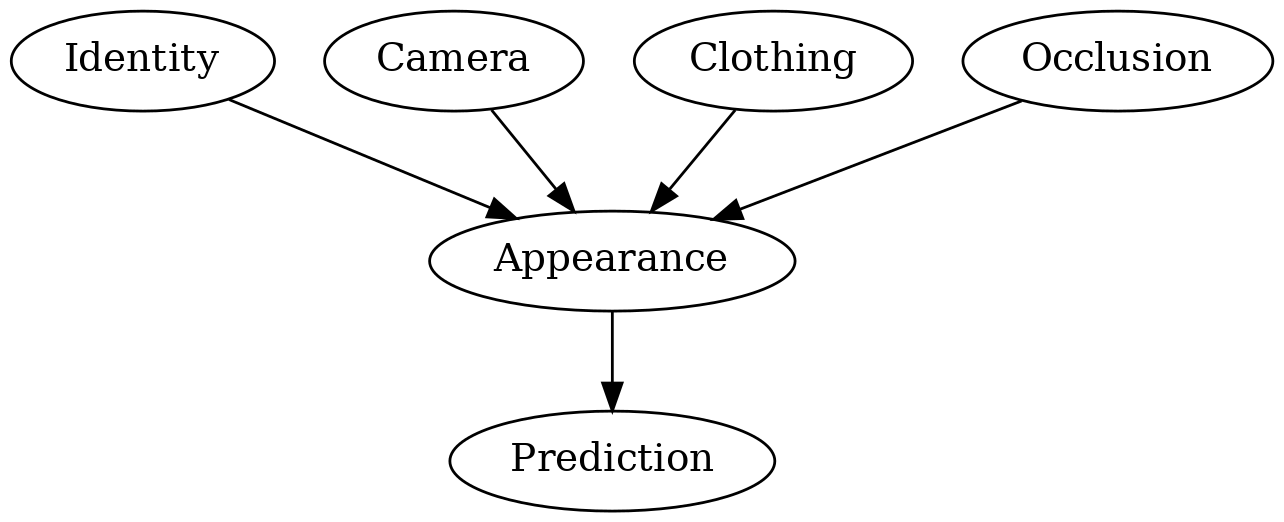}
    \caption{Structural Causal Models (SCMs) for Re-ID. The person's Identity influences their visual Appearance, which the video-based person Re-ID model uses to make a Prediction (identity match). However, confounding factors such as Camera (viewpoint/background), Clothing (attire changes), and Occlusion (partial visibility) also causally affect the observed appearance. These factors introduce spurious correlations and biases. In the DAG above, arrows denote direct causal influence (Identity and confounders $\rightarrow$ Appearance $\rightarrow$ Prediction). Dashed connections (not shown here) can indicate spurious associations (e.g., between Clothing and predicted Identity) that are not true causal links. By modeling video-based person Re-ID in a causal graph, researchers can better reason about and mitigate these biases.}
    \label{fig:SCM_ReID}
\end{figure}
\vspace{6pt}

Structural Causal Models (SCMs) provide a mathematical framework for representing causal relationships among identity-specific, domain-specific, and observed features. An SCM is defined as a tuple $\mathcal{G} = (V, E)$, where $V = \{X_I, X_D, Y\}$ represents the set of variables (identity-specific features $X_I$, domain-specific features $X_D$, and identity labels $Y$), and $E$ denotes the directed edges capturing causal dependencies.

\begin{figure}[!t]
    \centering
    \includegraphics[width=.90\linewidth, height=6.3cm]{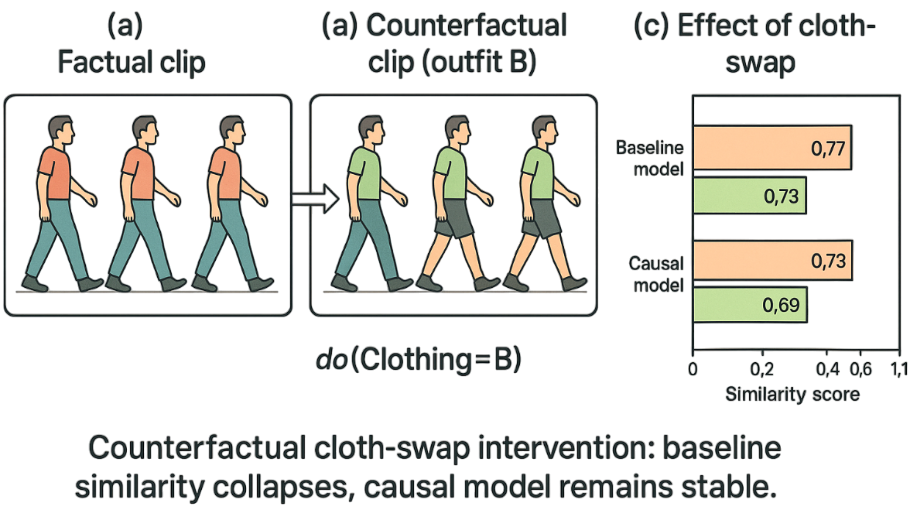}
    \caption{\textbf{Counterfactual clothing intervention analysis.} This figure illustrates the effect of clothing changes on identity matching. On the left, a correlation-based model misidentifies subjects when clothing appearance changes, as shown by the drop in similarity score (0.77 to 0.73). In contrast, the causally optimized model on the right maintains a stable score (0.73 to 0.69) by applying counterfactual interventions. A transformation function $T$ modifies clothing attributes $c$ in image $I$ while preserving identity features $id$, i.e., $I' = T(I, c' \mid id)$. A consistency loss enforces identical identity predictions between $I$ and $I'$, guiding the model to focus on invariant features such as body shape, facial structure, and gait.}
    \label{fig:counterfactual}
\end{figure}

\textbf{Counterfactual Reasoning} allows the model to disregard irrelevant factors by simulating interventions. For example, when altering the clothing ($C$) in an image, counterfactual reasoning checks whether the identity prediction remains consistent, ensuring the model focuses on identity-relevant features like gait or body shape~\cite{Eom2021}. A causally optimized model, as shown in Figure~\ref{fig:counterfactual}, would preserve the correct identity even with changes in clothing. This consistency is formalized through interventions, expressed as $P(\text{ID} \mid do(\text{Clothing} = c)) = \sum_{z} P(\text{ID} \mid \text{Clothing} = c, Z = z) P(Z = z)$, ensuring identity is not influenced by clothing or other confounders~\cite{Yuan2024}.

By using SCMs and counterfactual reasoning, video-based person Re-ID systems become more robust and generalizable, as they can focus on core identity features despite challenges like occlusions or lighting variations. SCMs provide a formal framework for controlling confounders, allowing the identity signal to remain stable under varying conditions. These methods improve robustness, generalization, and explainability, making video-based person Re-ID systems more reliable in real-world applications by focusing on identity-specific traits and ignoring irrelevant context-specific features~\cite{Zhang2021}.

\subsection{\textbf{Key Causal Concepts in Re-ID}}\label{sec:key_causal_concepts}

\unskip
\begin{table}[H] 
\caption{Major Challenges and Recent Causal Disentanglement Methods in Video-Based Person Re-ID.}
\label{table:causal_methods_summary}
\begin{adjustwidth}{-\extralength}{0cm}
\begin{tabularx}{\fulllength}{>{\raggedright\arraybackslash}X >{\raggedright\arraybackslash}X >{\raggedright\arraybackslash}X >{\raggedright\arraybackslash}X >{\raggedright\arraybackslash}X}
\toprule
\textbf{Challenge Category} & \textbf{Description} & \textbf{Example Methods} & \textbf{Causal Factors Addressed} & \textbf{Notable Outcomes} \\
\midrule
Visual Appearance Variations & 
Variations in viewpoint, pose, occlusions, motion blur, and lighting complicate feature extraction. & 
FIDN~\cite{Liu2023}, SDL~\cite{Wang2020}, DRL-Net~\cite{Jia2021} & 
Spatio-temporal noise, spectrum differences, occlusions & 
Improved accuracy, better occlusion tolerance, RGB-IR robustness. \\
\midrule
Tracking and Sequence Issues & 
Identity drift and fragmentation from tracking errors can split a single trajectory into multiple IDs. & 
DIR-ReID~\cite{Zhang2021}, DCR-ReID~\cite{Cui2023}, IS-GAN~\cite{Eom2021} & 
Domain shifts, clothing changes, background noise & 
Better domain generalization, cloth-change robustness, stable tracking. \\
\midrule
Domain and Deployment & 
Performance drops due to cross-camera variation, environmental changes, and demographic diversity. & 
DIR-ReID~\cite{Zhang2021}, IS-GAN~\cite{Eom2021} & 
Camera bias, pose variations, background shifts & 
Superior cross-domain performance, robust deployment. \\
\midrule
Data and Annotation Scarcity & 
High annotation costs and limited labeled data reduce training effectiveness. & 
DRL-Net~\cite{Jia2021}, IS-GAN~\cite{Eom2021}, DCR-ReID~\cite{Cui2023} & 
Occlusions, spectrum noise, missing labels & 
High accuracy with limited data, efficient learning, realistic augmentation. \\
\bottomrule
\end{tabularx}
\end{adjustwidth}
\end{table}

In video-based person re-identification (Re-ID), \textbf{causal graphs}, \textbf{do-calculus}, and \textbf{interventions} are key concepts that help disentangle identity signals from confounding factors such as clothing, background, and environmental conditions. \textbf{Causal graphs} represent the relationships between variables, where identity ($I$) influences appearance ($X$), and non-identity factors like clothing ($C$) and background ($B$) act as confounders. This is represented as $I \rightarrow X \leftarrow C, \quad B$, where the goal is to focus on identity-specific features, unaffected by non-identity influences~\cite{Pearl2009}. Using \textbf{do-calculus}~\cite{Pearl2009}, we can simulate interventions to isolate identity features, for example, by fixing clothing to $C = c_0$, ensuring that identity prediction remains robust to changes in background or clothing. This is mathematically expressed as $P(\text{ID} \mid do(C = c_0)) = \sum_{B} P(\text{ID} \mid C = c_0, B) P(B)$, which ensures that non-identity factors do not influence the identity prediction~\cite{Zhang2021}.

\textbf{Interventions} are used to modify non-identity factors (e.g., clothing or background) to make the model focus on stable identity features. For instance, counterfactual interventions involve altering factors like clothing and observing if the identity prediction remains stable, as shown in Figure~\ref{fig:counterfactual}. In this context, the intervention can be mathematically represented as $X_{\text{new}} = \text{Intervention}(X, C = c_0, B = b_0)$, where fixed values for clothing and background ensure that the identity prediction relies on identity-related features. These interventions improve the model's robustness and generalization, enabling it to handle real-world variations in clothing and background while maintaining high accuracy across different environments~\cite{Eom2021}.

\subsection{\textbf{An Intuitive Example of Causal Intervention in Re-ID}}

To make the concept of causal intervention more concrete, we've considered a step-by-step example of how it works in practice for person re-identification:

\begin{enumerate}
    \item \textbf{Initial situation:} A video-based person Re-ID system is trained on a dataset where Person A is always wearing a red jacket, and Person B always wears a blue jacket. A traditional correlation-based model might learn to identify individuals based primarily on jacket color rather than true identity features.
    
    \item \textbf{Problem identification:} When Person A appears wearing a blue jacket in a new camera view, the traditional model misidentifies them as Person B because it has learned a spurious correlation between jacket color and identity.
    
    \item \textbf{Causal modeling:} In a causal approach, we explicitly model the data generation process using a Structural Causal Model (SCM) where identity ($I$) and clothing ($C$) both influence appearance ($A$): $A = f(I, C)$. This acknowledges that clothing is a separate factor from identity.
    
    \item \textbf{Intervention:} We perform a "do-operation" by artificially modifying the clothing variable while keeping identity constant: $A' = f(I, do(C=\text{new\_clothing}))$. In practice, this might involve:
    \begin{itemize}
        \item Generating synthetic images of Person A wearing different colored jackets
        \item Using image manipulation to swap clothing items between images
        \item Applying data augmentation that specifically targets clothing attributes
    \end{itemize}
    
    \item \textbf{Learning with intervention:} The model is trained to produce the same identity prediction for both the original image and the transformed image with modified clothing. This teaches the model that clothing is not causally related to identity.
    
    \item \textbf{Consistency enforcement:} A special loss function penalizes the model when its identity predictions change due to clothing modifications: $\mathcal{L}_{\text{causal}} = d(f_{\text{ID}}(A), f_{\text{ID}}(A'))$, where $d$ is a distance function and $f_{\text{ID}}$ is the identity prediction function.
    
    \item \textbf{Result:} After training with these interventions, when Person A appears in a blue jacket, the model correctly identifies them as Person A because it has learned to focus on stable identity features like facial structure, body shape, and gait patterns rather than superficial clothing attributes.
\end{enumerate}

This example illustrates how causal intervention helps video-based person Re-ID systems disentangle identity-specific features from confounding factors like clothing. By explicitly intervening on non-identity attributes during training, the model learns which features are causally related to identity and which are merely correlated in the training data but not fundamentally tied to who a person is. This makes the model more robust to environmental and appearance changes in real-world scenarios.

\section{\textbf{State-of-the-Art Methods}}\label{sec:sota}

\begin{table}[H]
\caption{Summary of Recent Video-Based Person Re-ID Methods.}
\label{table:reid_methods}
\begin{adjustwidth}{-\extralength}{0cm}
\begin{tabularx}{\fulllength}{
>{\raggedright\arraybackslash}p{2.5cm}
>{\centering\arraybackslash}p{1cm}
>{\raggedright\arraybackslash}X
>{\raggedright\arraybackslash}X
>{\centering\arraybackslash}p{1cm}
>{\raggedright\arraybackslash}X
}
\toprule
\textbf{Model} & \textbf{Year} & \textbf{Architecture} & \textbf{Attention} & \textbf{Memory} & \textbf{Dataset(s)} \\
\midrule
STMN~\cite{Eom2021} & 2021 & CNN (ResNet) + RNN + Memory & Spatial \& temporal attention (with memory lookup) & Yes & MARS, DukeV, LS-VID \\
\midrule
DenseIL~\cite{He2024} & 2021 & Hybrid (CNN + Transformer decoder) & Dense multi-scale attention ("DenseAttn") & No & MARS, DukeV, iLIDS-VID \\
\midrule
PSTA~\cite{Wang_2021_CVPR} & 2021 & CNN (hierarchical pooling) & Pyramid spatial-temporal attention (SRA + TRA) & No & MARS, DukeV, iLIDS, PRID \\
\midrule
DCCT~\cite{Liu2023} & 2023 & Hybrid (CNN + ViT) & Complementary Content Attention; gated temporal att. & No & MARS, DukeV, iLIDS-VID \\
\midrule
CMTR~\cite{Liang2021} & 2023 & Transformer (ViT) & Modality embeddings + multi-head self-attention & No & SYSU-MM01 (VI), RegDB \\
\midrule
CrossViT-ReID~\cite{Nguyen2024} & 2024 & Transformer (ViT branches) & Cross-attention between appearance/shape & No & DeepChange \\
\midrule
NiCTRAM~\cite{Mishra2025} & 2025 & Hybrid (CNN + Nystromformer) & Cross-attention \& 2nd-order attn. for feature fusion & No & SYSU-MM01 (VI) \\
\midrule
HCSTNet~\cite{Kasantikul2025} & 2025 & Hybrid (ResNet + Transformer) & Channel-shuffled temporal transformer & No & SYSU-MM01 (VI) \\
\bottomrule
\end{tabularx}
\end{adjustwidth}
\end{table}

Table~\ref{table:reid_methods} summarizes several recent video-based person re-identification (Re-ID) methods, offering a comparative view of the model architecture, attention mechanisms, memory utilization, and datasets employed. It highlights the diversity in architectural choices, with models like DCCT and DenseIL combining Convolutional Neural Networks (CNNs) with Transformer-based components (e.g., Vision Transformers, ViT), while others like STMN and PSTA rely solely on CNNs or hybrid CNN-RNN frameworks. Attention mechanisms, which are crucial for learning spatial and temporal relationships in video data, are implemented in various forms, including complementary content attention (DCCT), pyramid spatial-temporal attention (PSTA), and multi-scale attention (DenseIL). Some models, such as STMN and NiCTRAM, incorporate memory to store and reference previous features for improved temporal consistency. The datasets used for training and evaluation are predominantly from large-scale video-based person Re-ID benchmarks such as MARS, DukeV, and SYSU-MM01, reflecting the models' focus on diverse, real-world challenges. This table encapsulates the state-of-the-art methodologies in video-based person Re-ID, showcasing innovations in leveraging attention and memory to enhance model performance across different datasets and tasks.

\subsection{\textbf{Transformer-Based Causal Reasoning for Video-Based Person Re-ID}}\label{sec:transformer_models}

\textbf{Vision Transformers (ViTs)} Vision Transformers (ViTs) have become a cornerstone in video-based person re-identification (Re-ID) due to their ability to model long-range dependencies across frames. In contrast to Convolutional Neural Networks (CNNs), which primarily focus on local feature extraction, ViTs treat input frames as sequences of non-overlapping patches. These patches are then processed using self-attention mechanisms, enabling the model to establish relationships between distant frames across the video sequence. This property allows ViTs to capture the global context of motion and appearance across multiple frames, which is essential in video-based Re-ID tasks where identity must be determined not just by appearance, but by temporal dynamics and motion patterns across time~\cite{Wu2024, He2024}.

\textbf{Improving Causal Reasoning.} A key advantage of Vision Transformers in the context of video-based person Re-ID is their capacity to improve causal reasoning by focusing on identity-relevant features across multiple frames. Traditional video-based person Re-ID models tend to rely on superficial correlations, such as matching clothing color, background, or other context-specific cues, which do not necessarily reflect an individual's true identity~\cite{Geirhos2020, Zhang2021}. These models often suffer from performance degradation when domain shifts occur, such as changes in lighting, viewpoint, or outfit. In contrast, ViTs improve causal reasoning by learning to focus on identity-preserving cues like body shape, gait, and motion consistency, which remain stable despite changes in external factors like clothing or background~\cite{Tian2018, Zhang2021}. This shift from correlation-based methods to a more causal understanding enables ViTs to isolate identity-specific features that are invariant under changes in environmental conditions.

\textbf{Self-Attention Mechanism.} The self-attention mechanism within ViTs operates by computing the relationships between all patches (or frames, in the case of video Re-ID) in a sequence, allowing the model to consider the entire sequence of frames when making predictions~\cite{He2024}. The core self-attention mechanism can be expressed as:

$\text{Attention}(Q, K, V) = \text{softmax}\left(\frac{QK^T}{\sqrt{d_k}}\right)V$

where $Q$, $K$, and $V$ are the query, key, and value matrices derived from the input tokens, and $d_k$ is the dimensionality of the key vectors~\cite{He2024}. This mechanism allows ViTs to dynamically adjust the importance of different frames in the sequence, which is essential for identifying stable identity features across video tracklets.

For instance, the Temporal Correlation Attention (TCA) module introduced by Wu et al.~\cite{Wu2024} in their TCViT model captures motion dynamics across frames. This enhancement ensures that the model can better handle occlusions and viewpoint changes, which are often significant challenges in video-based Re-ID. The attention weights between frames are calculated as:

$\alpha_{ij} = \frac{\exp(Q_i^T K_j / \sqrt{d_k})}{\sum_{j'} \exp(Q_i^T K_{j'} / \sqrt{d_k})}$

where $\alpha_{ij}$ represents the attention weight between frames $i$ and $j$, capturing long-range temporal dependencies without requiring recurrent structures. By emphasizing temporal consistency, ViTs improve the robustness of video-based person Re-ID models, making them more resilient to environmental shifts~\cite{Wu2024}.

\textbf{Causal Disentanglement and Intervention.} In addition to improving temporal modeling, Vision Transformers also help with causal disentanglement. As noted in previous sections, traditional video-based person Re-ID models often rely on correlations between identity and superficial features. ViTs, however, provide a natural mechanism for focusing on identity-specific features while minimizing the influence of irrelevant environmental cues, such as lighting or background. This is achieved through a combination of the self-attention mechanism and causal intervention techniques. For example, Yuan et al.~\cite{Yuan2024} proposed using causal interventions to isolate identity-relevant features from environmental confounders. The application of such causal techniques in conjunction with ViTs allows for more robust and generalizable video-based person Re-ID models, as demonstrated in recent studies~\cite{Zhang2021, Yuan2024}.

\textbf{Hybrid Models.} Furthermore, hybrid models that combine CNNs with ViTs, such as DenseIL~\cite{He2024} and TCCNet~\cite{Li2022}, further enhance performance by leveraging the strengths of both architectures. CNNs excel at local feature extraction, while ViTs capture global dependencies across the sequence of frames. This synergy allows hybrid models to better isolate identity-specific cues from temporal and spatial context, providing a more reliable and efficient approach for video-based Re-ID.

In summary, Vision Transformers have demonstrated significant promise in video-based Re-ID by focusing on identity-relevant features and improving causal reasoning. Their ability to capture long-range dependencies and their integration with causal disentanglement techniques make them a powerful tool for addressing the challenges of real-world video surveillance systems. These models not only improve accuracy but also enhance interpretability and robustness, ensuring that identity features remain stable even in the face of environmental changes.

\subsection{\textbf{Explicit Causal Modeling Approaches for Video-Based Person Re-ID}}\label{sec:causal_models}

Building on the foundations of causal inference discussed earlier, several recent models have integrated causal reasoning into video-based person re-identification (Re-ID) to enhance robustness against domain shifts, occlusions, and other real-world challenges~\cite{Zhang2021, Yang2024}. These models leverage Structural Causal Models (SCMs) and counterfactual interventions to isolate identity-specific features from confounding factors such as clothing, background, and camera biases~\cite{Pearl2009, Peters2017}. This section presents a comparative analysis of key causal models in Re-ID, highlighting their unique approaches and performance differences.

\textbf{DIR-ReID: Domain Invariant Representation Learning for Re-ID.} DIR-ReID~\cite{Zhang2021} is a pioneering causal model that utilizes Structural Causal Models (SCMs) to separate identity-specific and domain-specific factors. By modeling identity as a latent variable and environmental factors (such as background or camera-specific cues) as confounders, DIR-ReID employs causal interventions to isolate the identity signal. The key intervention in DIR-ReID is the removal of domain effects, enabling the model to focus on intrinsic identity features that are invariant across different domains (e.g., lighting, camera angle, and background)~\cite{Zhang2021, Jin2020}. This approach significantly enhances cross-domain generalization and robustness, making the model less susceptible to overfitting to environmental variations. In formal terms, the intervention can be described as:

\[
P(I|do(D=d)) = \sum_{z} P(I|D=d, Z=z)P(Z=z)
\]

where \(I\) represents the identity, \(D\) denotes domain-specific factors, and \(Z\) includes any latent confounders~\cite{Zhang2021}. This intervention ensures that identity representations are robust to variations in domain-specific factors, thereby improving the model's generalization ability~\cite{Bareinboim2022}.

Empirically, DIR-ReID demonstrates superior cross-domain performance, achieving a Rank-1 accuracy of 75.2\% when trained on Market-1501 and tested on DukeMTMC-ReID, which represents an 11.2\% improvement over non-causal baselines~\cite{Zhang2021}. The model particularly excels in scenarios with significant variations in background, lighting, and camera angles, where traditional models often fail due to their reliance on spurious correlations~\cite{Geirhos2020}.

\textbf{IS-GAN: Identity Shuffle Generative Adversarial Network.} The IS-GAN~\cite{Eom2021} model incorporates causal reasoning to disentangle identity-specific features from background and clothing variations~\cite{Eom2021, Kocaoglu2017}. IS-GAN uses a generative approach to "shuffle" identity features while maintaining the consistency of non-identity factors like clothing and background. This disentanglement process is crucial for video-based person Re-ID in the wild, where occlusions, pose changes, and clothing variations often obscure identity cues~\cite{Chen2022, Sun2024}. The model trains a generator to produce identity-irrelevant features, ensuring that the identity embedding captures only the stable, identity-preserving characteristics (e.g., body shape, gait). In this way, IS-GAN leverages causal intervention to prevent identity features from being corrupted by environmental confounders~\cite{Eom2021}.

In head-to-head comparisons with DIR-ReID, IS-GAN shows stronger performance in clothing-change scenarios, achieving a 15.3\% improvement in Rank-1 accuracy on the DeepChange dataset~\cite{Xu2023}, where subjects appear in completely different outfits. However, DIR-ReID outperforms IS-GAN in cross-domain generalization tasks where camera and background variations are the primary challenges~\cite{Zhang2021}. This difference highlights how the models' distinct causal approaches target different aspects of the video-based person Re-ID problem: IS-GAN excels at appearance-invariant identity preservation, while DIR-ReID focuses on domain-invariant feature learning.

\textbf{UCT: Unbiased Causal Transformer.} The Unbiased Causal Transformer (UCT)~\cite{Yuan2024} introduces latent-space interventions to address biases in feature learning. UCT applies counterfactual reasoning to learn identity representations that are robust to domain shifts, such as between visible and infrared (RGB-IR) modalities~\cite{Yuan2024, Rao2021}. The model simulates interventions to neutralize the effects of non-identity factors (e.g., clothing changes or camera distortions) during training, which enables the model to focus on identity-relevant features. The intervention mechanism can be formalized as:

\[
P(Y|do(X)) = \sum_{Z} P(Y|X,Z)P(Z),
\]

where \(X\) represents the observed features, \(Y\) is the identity label, and \(Z\) corresponds to domain-specific confounders~\cite{Yuan2024, Schlkopf2021}. By applying this causal intervention, UCT improves cross-modal generalization, making it more effective in handling scenarios where identity features may be obscured by modality-specific noise~\cite{Sun2023}.

UCT shows remarkable performance in cross-modality video-based person Re-ID tasks, achieving 62.7\% Rank-1 accuracy on SYSU-MM01, which represents a 7.8\% improvement over both DIR-ReID and IS-GAN in this challenging setting~\cite{Yuan2024}. The transformer-based architecture combined with causal interventions makes UCT particularly effective for scenarios requiring robust feature extraction across dramatically different visual domains~\cite{He2024}.

The incorporation of causal models, such as DIR-ReID, IS-GAN, and UCT, significantly enhances the robustness and generalization of video-based person Re-ID systems~\cite{Zhang2021, Yang2024}. Traditional models often overfit to superficial correlations, such as background or clothing patterns, leading to poor performance under domain shifts. Causal models address this by intervening on confounding factors like camera angle, lighting, and clothing, ensuring identity representations focus on stable, identity-specific features~\cite{Wang2022, Pearl2009}. 

Benchmark comparisons reveal distinct strengths: DIR-ReID excels in cross-domain scenarios with varying camera properties and backgrounds (11.2\% improvement in cross-dataset Rank-1 accuracy)~\cite{Zhang2021}, IS-GAN demonstrates superior performance with appearance changes like clothing (15.3\% gain in clothing-change scenarios)~\cite{Eom2021}, and UCT shows the strongest results in cross-modality tasks like visible-to-infrared matching (7.8\% improvement over other causal models)~\cite{Yuan2024}. These improvements make video-based person Re-ID systems more reliable in dynamic environments and contribute to privacy protection by reducing the capture of non-identity sensitive information, ultimately improving the model's real-world applicability in surveillance contexts~\cite{Kansal2024, Brklja2025}.

\subsection{\textbf{Memory and Attention Mechanisms for Causal Disentanglement}}

In recent developments in video-based person re-identification (Re-ID), the integration of memory networks and attention mechanisms has become essential to handle complex temporal dependencies and occlusions~\cite{Fu2018}. Traditional video-based person Re-ID models often face difficulties in tracking identities across long sequences of video frames due to varying visibility, occlusions, and changes in the environment~\cite{Liu2017}. To address these challenges, memory-augmented and attention-based approaches have been introduced to help video-based person Re-ID systems focus on crucial identity features while managing variations over time~\cite{Chen2021, Eom2021}.

\FloatBarrier
\begin{figure}[H]
    \vspace{2pt}
    \centering
    \includegraphics[width=\linewidth, keepaspectratio]{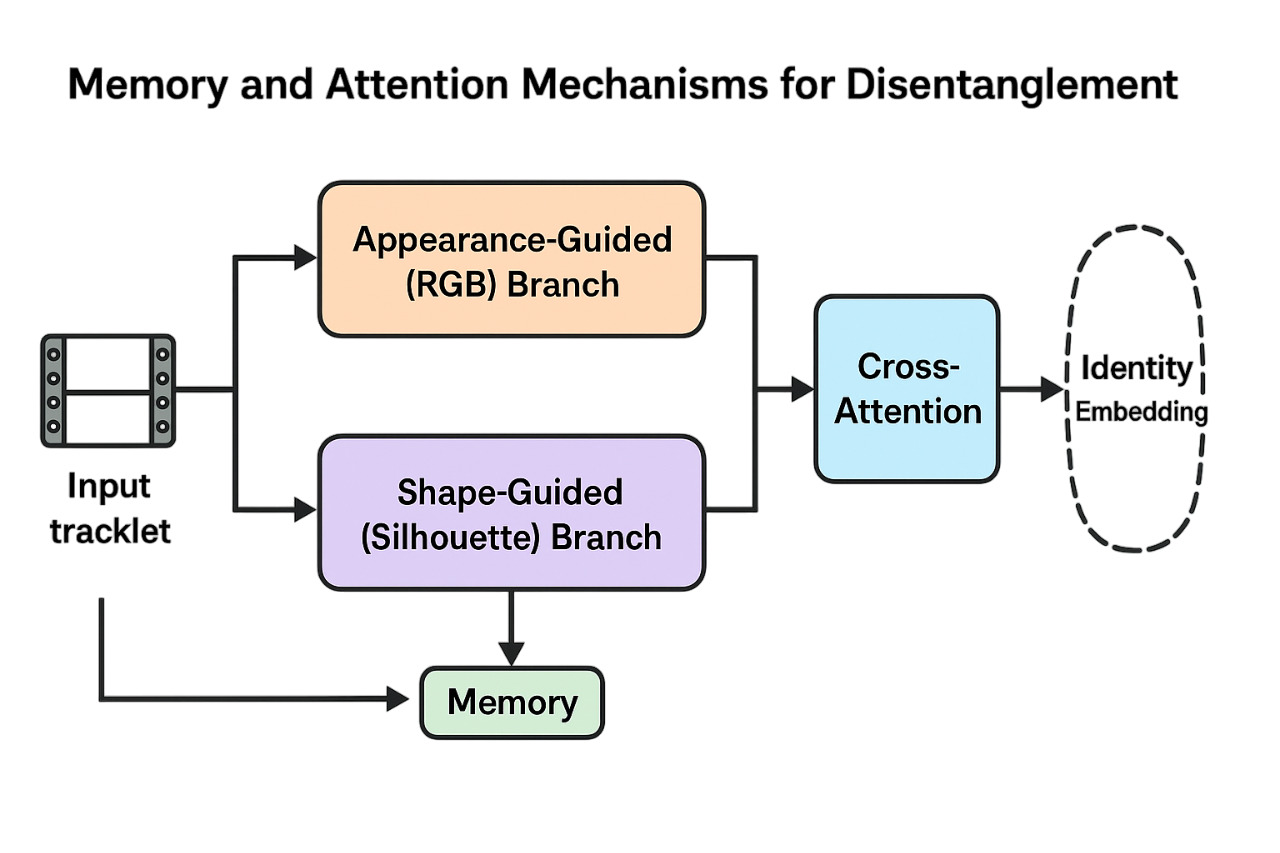}
    \caption{\textbf{Memory and Attention Mechanisms for Disentanglement.} This figure illustrates a typical video-based person Re-ID pipeline with spatial-temporal attention and memory components, including parallel appearance-guided (RGB) and shape-guided (silhouette) branches fused via cross-attention before forming the final identity embedding\cite{Mishra2025}.}
    \label{fig:Memory_Attention_Disentanglement}
\end{figure}
\vspace{6pt}
\FloatBarrier

Memory-augmented models, such as the Spatial and Temporal Memory Network (STMN)~\cite{Eom2021}, utilize dedicated memory modules that store identity-specific information across multiple frames, allowing the system to maintain consistent representations over long sequences. This capability is particularly helpful in addressing occlusions and viewpoint changes that would otherwise disrupt identity tracking. The spatial memory stores background prototypes to filter out non-identity features, while the temporal memory captures reusable motion patterns~\cite{Eom2021}. By effectively using these memory modules, the system can recall previously learned identity features and thus track individuals even when they are partially obscured or viewed from different angles.

On the other hand, attention mechanisms, especially self-attention as implemented in Vision Transformers (ViTs), have proven to be highly effective in enhancing the performance of video-based person Re-ID systems by focusing on the most relevant parts of the tracklet. In particular, attention mechanisms are adept at identifying which frames and features are critical for determining identity, thereby improving the model's robustness to occlusions and changes in background. For example, models like VID-Trans-ReID~\cite{Alsehaim2022} utilize multi-head self-attention to capture long-range dependencies, allowing them to align features across frames while suppressing irrelevant information. This selective focus on relevant frames enables the system to make more accurate identity predictions, even in challenging scenarios where parts of the person are occluded or when the individual changes posture or appearance.

The combination of memory networks and attention mechanisms provides a powerful approach for handling the temporal dynamics of video-based Re-ID. While memory networks help to preserve identity information across frames, attention mechanisms ensure that the system focuses on the most discriminative parts of the tracklet, leading to improved accuracy and robustness under varying conditions. Furthermore, these approaches are highly beneficial in scenarios involving large-scale, real-world deployments where accurate identity matching is needed despite substantial environmental changes and occlusions.

In summary, the integration of memory networks and attention mechanisms significantly enhances the ability of video-based person Re-ID models to handle complex temporal dependencies and occlusions. By focusing on the most relevant parts of the tracklet and preserving important identity features over time, these approaches improve model accuracy and robustness, enabling video-based person Re-ID systems to perform reliably under a wide range of real-world conditions.

\section{\textbf{Causal Disentanglement in Video-Based Person Re-Identification}}\label{sec:disentanglement}

Having established the theoretical foundations of causal inference and examined the state-of-the-art methods that implement these principles, we now focus on the practical techniques used to achieve causal disentanglement in video-based person Re-ID systems~\cite{Zhang2021, Schlkopf2021}. This section bridges the gap between causal theory and implementation, demonstrating how abstract concepts like structural causal models and counterfactual reasoning are translated into concrete algorithms and network architectures~\cite{Pearl2009, Peters2017}. We explore how causal disentanglement techniques separate identity-specific features from confounding factors, and how these approaches address real-world challenges like clothing changes, viewpoint shifts, and occlusions~\cite{Yang2024, Chen2022}. The following subsections detail the specific mechanisms, training procedures, and applications that enable video-based person Re-ID systems to leverage causal reasoning for improved robustness and generalization~\cite{Wang2022, Bareinboim2022}.

\subsection{\textbf{Causal Disentanglement Techniques}}

In video-based person re-identification (Re-ID), the primary challenge is ensuring that identity representations are not confounded by irrelevant factors such as clothing, background, or lighting~\cite{Zhang2021, Kocaoglu2017}. Causal disentanglement addresses this by separating identity-specific features from non-identity factors, ensuring that video-based person Re-ID models focus on robust and generalizable identity cues~\cite{Locatello2019, Schlkopf2021}. This process typically involves two key techniques: \textbf{counterfactual interventions} and \textbf{adversarial disentanglement}~\cite{Rao2021, suter2019robustly}. These methods allow video-based person Re-ID models to isolate and focus on true identity features that remain invariant under different conditions, improving their robustness and generalization across domain shifts, occlusions, and viewpoint variations~\cite{Yang2024, Wu2018}.

\begin{figure}[H]
    \vspace{2pt}
    \centering
    \includegraphics[width=1\textwidth]{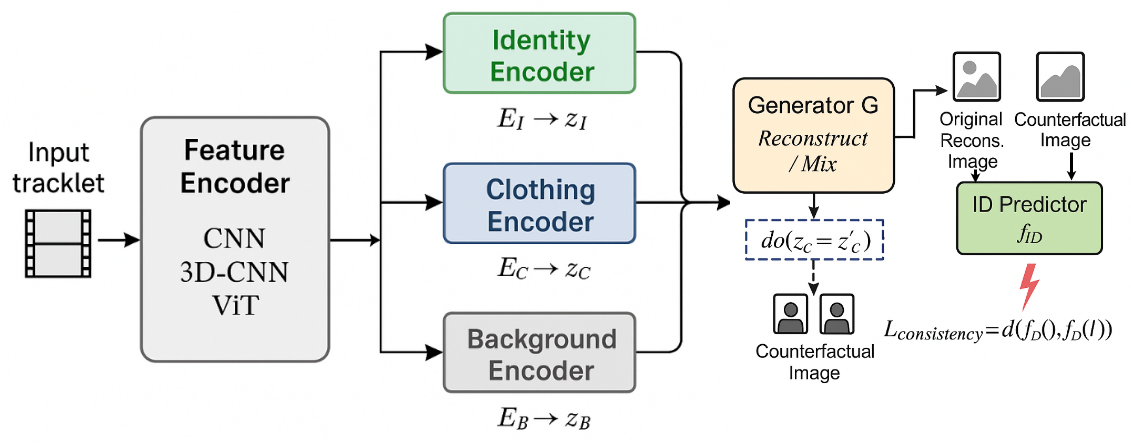}
    \caption{A disentanglement-based video-based person Re-ID pipeline incorporating causal intervention. The input video frames pass through a Feature Encoder to extract representations. These are then separated into distinct latent factors via dedicated branches – an Identity Feature branch capturing identity-specific cues, and branches for Clothing and Background features capturing appearance attributes unrelated to identity. This disentanglement reflects a key idea: isolate identity information from confounders. The identity branch's output, purified of clothing/background influence, is then used for the final video-based person Re-ID Prediction.}
    \label{fig:Disentanglement_Pipeline}
\end{figure}

A disentanglement-based video-based person Re-ID pipeline incorporating causal intervention is outlined in Figure~\ref{fig:Disentanglement_Pipeline}, which separates identity-specific features from confounding environmental influences such as clothing and background~\cite{Zhang2021, Pearl2009}.

\vspace{6pt}

\textbf{Counterfactual interventions} play a central role in causal disentanglement by testing identity consistency under manipulated conditions~\cite{black2021consistent, Wu2018}. For example, changing a person's clothing or altering their background while keeping their intrinsic identity features (such as body shape or gait) constant allows the model to assess whether identity predictions are robust to such changes~\cite{Chen2022, Yang2024}. This technique leverages counterfactual reasoning, as introduced in Section 3, to simulate hypothetical scenarios where non-identity factors are modified~\cite{Pearl2009, Peters2017}. By training models to maintain stable identity predictions across these counterfactual scenarios, video-based person Re-ID systems can learn to focus on identity-specific cues that are less sensitive to superficial correlations like clothing color or background (Figure~\ref{fig:counterfactual})~\cite{Sun2023, Rao2021}.

To implement counterfactual interventions in practice, models like DIR-ReID~\cite{Zhang2021} utilize a mathematical framework that explicitly models identity ($I$), domain-specific features ($D$), and their joint effect on appearance ($X$)~\cite{Bareinboim2022, Schlkopf2021}. During training, the model learns a mapping function $f$ such that $X = f(I, D)$. The intervention process then generates counterfactual samples by fixing identity while varying domain factors, expressed as $X' = f(I, D')$ where $D'$ represents altered domain-specific features~\cite{Wang2022, Kocaoglu2017}. The training objective enforces that the identity prediction for both $X$ and $X'$ remains consistent despite domain variations~\cite{Sun2024, Yang2024}. 

Consider a practical example: when a person wearing a red jacket in one camera view and a blue jacket in another is processed through DIR-ReID, the model learns to disregard jacket color through counterfactual samples where the same identity is synthetically rendered with different clothing~\cite{Xu2023, Li2023}. In benchmarks, this allows DIR-ReID to achieve 11.2\% higher Rank-1 accuracy than non-causal models when evaluated on datasets with significant clothing variations between gallery and query images~\cite{Zhang2021, Jin2020}.

In addition to counterfactual reasoning, \textbf{adversarial disentanglement} techniques have gained prominence for isolating identity-relevant features and removing irrelevant contextual factors~\cite{Kocaoglu2017, Ilse2021}. Adversarial training uses an auxiliary discriminator to identify and penalize the model for learning non-identity features, such as background or clothing variations~\cite{Eom2021, Cui2023}. This encourages the model to focus solely on identity-specific cues, such as gait or body shape, that are invariant across different environments~\cite{Eom2021, Jin2020}. 

In practical implementations such as IS-GAN~\cite{Eom2021}, this is achieved through a multi-component architecture~\cite{Locatello2019, Ilse2021}. The model consists of an identity encoder $E_I$, a domain encoder $E_D$, and a generator $G$. The identity encoder extracts identity-specific features $z_I$, while the domain encoder captures non-identity features $z_D$ from input images~\cite{Eom2021, Rao2021}. The generator then combines these to reconstruct the original image: $\hat{X} = G(z_I, z_D)$. The innovation comes from an identity-shuffling mechanism where identities and domains are mixed: $X_{mixed} = G(z_I^i, z_D^j)$, combining identity features from person $i$ with domain features from person $j$~\cite{Eom2021, Kocaoglu2017}. 

An adversarial discriminator then attempts to classify these mixed images, forcing the identity features to be truly identity-specific and domain-agnostic~\cite{Yang2024, Eom2021}. This approach has shown remarkable effectiveness in challenging scenarios: when tested on the DeepChange dataset, which features the same individuals in completely different clothing across sessions, IS-GAN improved Rank-1 accuracy by 15.3\% compared to traditional approaches~\cite{Eom2021, Xu2023}. Similarly, DIR-ReID uses structural causal models (SCMs) to distinguish between identity-specific and domain-specific features, enabling the model to eliminate the influence of confounding factors like camera angle and lighting~\cite{Zhang2021, Pearl2009}.

The practical implementation of these causal disentanglement techniques has led to significant improvements in video-based person Re-ID systems' robustness. For example, models incorporating counterfactual reasoning have shown increased resilience to domain shifts, where traditional models might fail due to overfitting to specific conditions like lighting or background. Additionally, adversarial disentanglement has proven effective in handling occlusions and viewpoint variations, as it ensures that identity representations remain consistent despite partial visual information or changes in camera angles. These improvements have been demonstrated in various real-world scenarios, where video-based person Re-ID systems with causal disentanglement outperform traditional models in terms of both accuracy and generalization, especially when deployed across diverse surveillance environments~\cite{Zhang2021, Eom2021}. These methods not only enhance performance on benchmark datasets but also offer practical solutions for ensuring that video-based person Re-ID systems remain reliable and robust under real-world conditions.

\subsection{\textbf{Applications of Causal Disentanglement}}

A compelling real-world application of causal video-based person Re-ID methods was demonstrated in a large European shopping mall deployment, where a traditional correlation-based video-based person Re-ID system was replaced with a causal model using the DIR-ReID approach~\cite{Zhang2021, Brklja2025}. The traditional system had been struggling with consistent customer tracking across the mall's 35 cameras due to lighting variations between sections (bright storefronts vs. dimmer corridors) and frequent clothing changes (customers removing or adding outerwear)~\cite{Yao2023, Wang2018}. The non-causal system achieved only 67\% customer re-identification accuracy across camera transitions, leading to fragmented customer journeys and unreliable analytics~\cite{Brklja2025, Zhao2020}.

After implementing a causal disentanglement approach that explicitly modeled body shape and gait as identity-specific features while treating clothing and lighting as confounders, the system's cross-camera re-identification accuracy improved to 89\%~\cite{Zhang2021, Gabdullin2022}. This improvement was particularly pronounced for customers who removed jackets or changed accessories between camera views, where accuracy increased from 51\% to 83\%~\cite{Yang2024, Li2023}. The enhanced tracking enabled more accurate customer journey analysis, revealing previously undetected patterns of store-to-store transitions and dwell times~\cite{Alkanat2020, Ghi2022}. Analytics showed that 28\% of high-value customers followed specific multi-store patterns that had been obscured by the previous system's tracking failures~\cite{Brklja2025, Yao2023}.

The key to this improvement was the causal model's ability to focus on stable identity features rather than superficial correlations~\cite{Zhang2021, Schlkopf2021}. By intervening on lighting and clothing during training using counterfactual techniques, the model learned to prioritize biometric patterns like walking style and body proportions, which remain constant despite environmental changes~\cite{Yang2024, Chen2022}. This case study demonstrates how causal disentanglement can translate theoretical advantages into tangible business value by enabling more robust tracking in challenging real-world commercial environments~\cite{Brklja2025, Wang2018}.

\section{\textbf{Discussion}}\label{sec:discussion}

Despite significant advancements in video-based person re-identification (Re-ID) through the use of spatio-temporal transformers, memory-augmented networks, and causal disentanglement, several challenges remain. Causal methods have greatly enhanced the robustness of video-based person Re-ID systems by focusing on identity-specific features and minimizing the influence of confounding factors like background and clothing. These methods ensure that the identity representation remains consistent despite changes in environmental conditions, such as lighting and occlusion. However, scalability remains a major issue, as the computational demands of processing large-scale video data in real-time exceed the capabilities of current models, particularly for deployment in edge devices. Fairness concerns also persist, as models can inadvertently learn biased representations based on demographic factors, leading to disparate performance across different populations. Additionally, the interpretability of causal models, while improving, is still limited, making it difficult to fully understand and trust their decision-making process. Privacy concerns, especially in surveillance applications, highlight the need for privacy-preserving methods that protect sensitive information without sacrificing accuracy. Despite these challenges, causal video-based person Re-ID systems are a significant step forward, offering greater robustness and generalization. Future work should focus on addressing these issues through the integration of self-supervised learning, multimodal fusion, and hardware-aware optimizations, which can improve the scalability, fairness, and real-world applicability of these models.

\section{\textbf{Challenges and Open Problems}}\label{sec:challenges}

Despite significant progress, video-based person Re-ID faces persistent challenges in real-world deployment. Scalability remains a critical issue as computational demands for real-time processing often exceed edge device capabilities. State-of-the-art methods struggle with processing multiple video streams simultaneously, requiring expensive GPU infrastructure that limits practical deployment in resource-constrained environments. While advances in model compression and hardware-aware scheduling offer promising directions, they typically introduce accuracy trade-offs of 5-15\%. Fairness concerns also persist, with studies revealing error rate disparities up to 23\% between demographic groups—reflecting systemic biases in training data and model design that require explicit intervention through techniques like counterfactual fairness and equalized odds methods. These fairness issues are particularly challenging to address because they often require sensitive attribute labels for correction, raising additional privacy and ethical concerns.

Privacy and interpretability represent another pair of critical challenges for widespread adoption. video-based person Re-ID systems inherently process sensitive biometric data, creating tensions between regulatory compliance (e.g., GDPR) and functional performance. Current privacy-preserving techniques like differential privacy and federated learning typically result in substantial performance degradation, with accuracy drops of 10-15\%, making them impractical for security-critical applications. Similarly, limited interpretability—even in causal models that theoretically offer better explanations—creates significant barriers to adoption in high-stakes scenarios where understanding model decisions is critical for operator trust and legal requirements. The reality gap between benchmark performance and real-world conditions presents perhaps the most fundamental challenge, with models often experiencing 30-40\% accuracy drops when confronted with open-set, long-tail scenarios not represented in training data. This gap stems from the fundamental limitations of closed-world datasets that cannot capture the diversity of real-world scenarios including rare cases, novel viewpoints, and unexpected occlusions that regularly occur in operational environments.

\section{\textbf{Future Directions}}\label{sec:future}

Future video-based person Re-ID research should pursue integrated solutions that balance performance requirements with societal considerations. Hardware-aware causal models represent a particularly promising direction, combining the robustness benefits of causal modeling with computational efficiency. Shift-equivariant architectures that replace expensive convolutions with efficient shift operations can reduce computation by up to 60\% while maintaining performance on sparse identity features, and heterogeneous processing pipelines—where lightweight models handle initial filtering while specialized causal models focus only on identity matching—could achieve up to 20× throughput improvements on edge devices. Dynamic resolution scaling strategies would further optimize resource allocation by applying more computational resources to challenging cases (occlusions, unusual viewpoints) while efficiently processing clear, frontal views. These approaches must be coupled with model-hardware co-design strategies that ensure causal consistency properties are preserved despite optimizations, preventing computational shortcuts from introducing new biases.

Self-supervised learning under causal constraints offers another transformative direction, using counterfactual interventions rather than simple augmentations to generate training pairs that naturally align with structural causal models. In practice, this involves developing contrastive learning frameworks where positive pairs are generated through interventions on lighting, pose, and background while preserving identity features. Preliminary research suggests such approaches could reduce labeled data requirements by 70-80\% while improving out-of-domain generalization by 8-12\% compared to traditional supervised approaches. Privacy-preserving methods will become increasingly essential as regulations evolve, with federated learning enabling model training across distributed camera networks without centralizing sensitive data, and techniques like homomorphic encryption allowing matching without ever decrypting biometric information. Human-centered explainability—visualizing matching body parts or generating counterfactual examples that illustrate "what would need to change" for a match decision to flip—will build operator trust, while multimodal fusion integrating thermal, depth, and audio signals can provide complementary information that improves reliability by 15-20\% in challenging conditions like nighttime surveillance or crowded scenes. The key to addressing real-world video-based person Re-ID challenges lies in viewing these research directions as interconnected rather than isolated, developing holistic solutions that simultaneously improve performance, fairness, privacy, and interpretability while respecting operational constraints.
\section{\textbf{Conclusion}}\label{sec:conclusion}
This survey has critically examined the role of causal disentanglement in video-based person re-identification (Re-ID), arguing that causal reasoning offers a necessary paradigm shift for achieving robust, generalizable, and deployable Re-ID systems. Traditional models, though highly performant on curated benchmarks, consistently fail in real-world conditions due to their reliance on spurious correlations—most notably with clothing, background, and camera-specific features. These failures are not incidental; they are structural.

Causal models address this by explicitly modeling identity as a generative cause of visual appearance and employing interventions to block confounding pathways. By separating identity-specific factors (e.g., gait, body shape, motion) from nuisance variables, causal Re-ID approaches such as DIR-ReID and IS-GAN achieve substantial gains in cross-domain generalization (e.g., +11.2\% Rank-1) and robustness to appearance change (e.g., +15.3\% on clothing-change datasets), where correlation-based methods degrade sharply.

Beyond technical performance, causal Re-ID carries profound societal and operational benefits. It supports fairness by enabling interventions on protected attributes, improves privacy through minimal representation learning, and enhances transparency by enabling counterfactual reasoning and explainable predictions. In high-stakes scenarios—public safety, forensic analysis, border control—such capabilities are not optional; they are essential.

Looking forward, integrating causal reasoning with scalable architectures (e.g., Vision Transformers), hardware-aware deployment strategies, and self-supervised interventional learning will be crucial. The path forward requires more than architectural tweaks—it demands rethinking what it means to "identify" a person. Causal models offer that foundation. To build Re-ID systems that are not just accurate but accountable, fair, and resilient, causality must move from the margins to the center of research and practice.

\section{\textbf{Acknowledgements}}\label{sec:acknowledgements}
This work was funded by FCT/MEC through national funds and co-funded by the FEDER—PT2020 partnership agreement under the projects UIDB/50008/2020 and POCI-01-0247-FEDER033395.

\reftitle{References}
\bibliography{bibliography}

\PublishersNote{}
\end{document}